\documentclass[letterpaper, 10 pt, conference]{ieeeconf}  

\IEEEoverridecommandlockouts                              

\usepackage{graphics} 
\usepackage{epsfig} 
\usepackage{times} 
\usepackage{amsmath} 
\usepackage{amssymb}  
\usepackage{biblatex}
\usepackage{multirow}
\usepackage{multicol}
\usepackage{booktabs}
\usepackage{colortbl}
\usepackage{hyperref}
\usepackage{enumerate}
\usepackage{color}
\usepackage{tabularx}
\usepackage{hyperref}
\usepackage{pifont}
\usepackage{algorithm}
\usepackage{algpseudocode}
\usepackage{amsmath}
\usepackage{adjustbox}
\usepackage{subfig}
\usepackage{diagbox}
\usepackage[usenames,dvipsnames,table]{xcolor}
\hypersetup{
    colorlinks=true,
    linkcolor=MidnightBlue,
    filecolor=magenta,      
    urlcolor=MidnightBlue,
    citecolor=MidnightBlue,
} 
\usepackage[font=small,labelfont=bf]{caption}

\newcommand\model{CAGE}
\newcommand\best[1]{\textbf{#1}}
\NewDocumentCommand\emojicage{}{
    \includegraphics[scale=0.03]{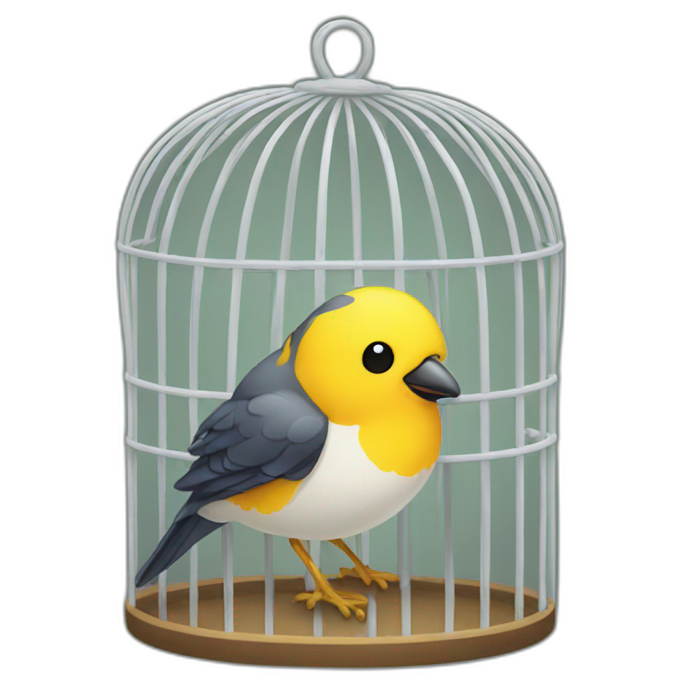}
}
\NewDocumentCommand\emojifreeze{}{
    \includegraphics[scale=0.015]{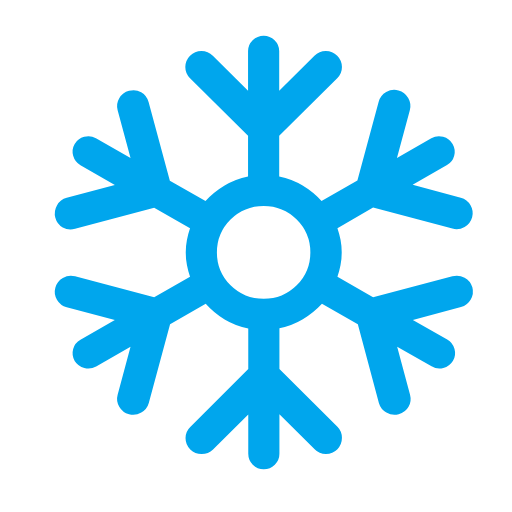}
}
\NewDocumentCommand\emojifire{}{
    \includegraphics[scale=0.015]{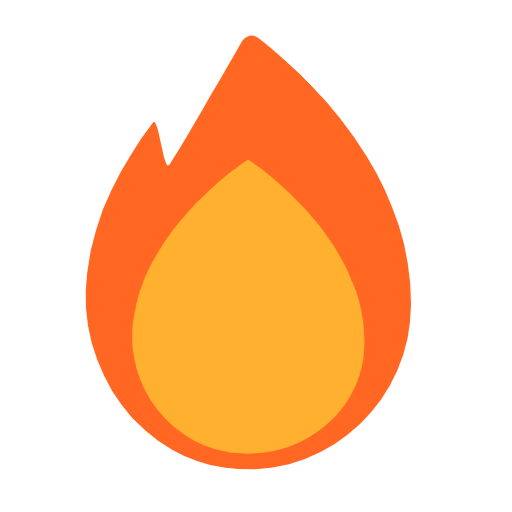}
}

\definecolor{darkgreen}{rgb}{0.01, 0.75, 0.24}

\title{\LARGE \bf
\model: \underline{C}ausal \underline{A}ttention Enables Data-Efficient\\ \underline{Ge}neralizable Robotic Manipulation
}

\author{Shangning Xia$^{1}$, Hongjie Fang$^{1,2}$, Cewu Lu$^{1,2,\dagger}$, Hao-Shu Fang$^{1,\dagger}$ 
\thanks{$^{1}$ Shanghai Jiao Tong University.} %
\thanks{$^{2}$ Shanghai Artificial Intelligence Laboratory.}
\thanks{$^\dagger$ Cewu Lu and Hao-Shu Fang are the corresponding authors.}
\thanks{\{ritchie\_xia, galaxies, lucewu\}@sjtu.edu.cn, fhaoshu@gmail.com.}
}

\begin{document}

\makeatletter
\let\@oldmaketitle\@maketitle
\renewcommand{\@maketitle}{\@oldmaketitle
\vspace{-0.1cm}
\centering
\includegraphics[width=0.9\linewidth]{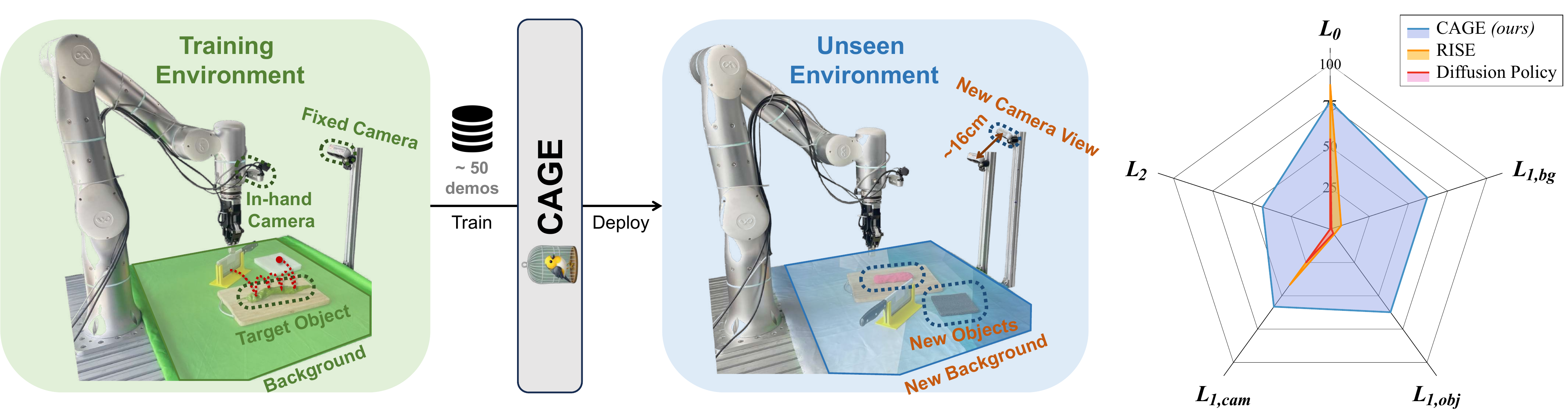}
\label{fig:teaser}
\vspace{-0.2cm}
\captionof{figure}{\textbf{\emojicage \model}~is a data-efficient generalizable robotic manipulation policy. With approximately 50 mono-distributed demonstrations, \model~can effectively complete the task in test environments with different levels of distribution shifts (See \S\ref{sec:gen-level} for details): training environment ($L_0$), similar environment ($L_1$), and unseen environment ($L_2$). Experiments demonstrate that \model~generalizes well to $L_1$ and $L_2$ environments and significantly outperforms prior works.}
\vspace{-0.2cm}
}%
\makeatother

\maketitle
\thispagestyle{empty}
\pagestyle{empty}
\addtocounter{figure}{-1}

\begin{abstract}
Generalization in robotic manipulation remains a critical challenge, particularly when scaling to new environments with limited demonstrations. This paper introduces \model, a novel robotic manipulation policy designed to overcome these generalization barriers by integrating a causal attention mechanism. \model~utilizes the powerful feature extraction capabilities of the vision foundation model DINOv2, combined with LoRA fine-tuning for robust environment understanding. The policy further employs a causal Perceiver for effective token compression and a diffusion-based action prediction head with attention mechanisms to enhance task-specific fine-grained conditioning. With as few as 50 demonstrations from a single training environment, \model~achieves robust generalization across diverse visual changes in objects, backgrounds, and viewpoints. Extensive experiments validate that \model~significantly outperforms existing state-of-the-art RGB/RGB-D approaches in various manipulation tasks, especially under large distribution shifts. In similar environments, \model~offers an average of 42\% increase in task completion rate. While all baselines fail to execute the task in unseen environments, \model~manages to obtain a 43\% completion rate and a 51\% success rate in average, making a huge step towards practical deployment of robots in real-world settings. Project website: \href{https://cage-policy.github.io/}{cage-policy.github.io}.
\end{abstract}

\section{Introduction}\label{sec:intro}

Generalization is one of the ultimate goals of robotic manipulation. Despite progress in certain areas like general grasping~\cite{anygrasp,dexgrasp}, broader manipulation tasks still require significant development. Recent advances in imitation learning~\cite{rt1, dp, rise, act} highlight the potential of behavior-cloning-based~\cite{bc} robotic manipulation policies in learning from demonstrations. However, behavior cloning often struggles with out-of-distribution scenarios, leading to compounded errors and task failures, which emphasizes the importance of generalization ability in the robotic manipulation policy.

Aiming at a generalizable robotic manipulation policy, previous research has concentrated on two main directions: (1) collecting or creating more demonstration data for policy training, and (2) searching for a better representation of environment perceptions for robotic manipulation tasks.

For the former direction, several studies have collected large-scale robotic datasets~\cite{rt1, rh20t, droid, oxe, bridge-v2} and trained robotic foundation models on those datasets. While these solutions do show an improved generalization performance in evaluations, a fundamental question remains: \textit{Does the policy itself truly generalize, or is the superior performance attributable to the policy having seen similar demonstrations during training?}

We believe that the upper limit of a policy's generalization ability is determined by its inherent structure, and merely increasing the amount of training data can only narrow the gap instead of elevating the limit. Therefore, we opt for the second path to \textit{address the generalization issue at its core}. Many studies~\cite{act3d, rise, dp3d, sgr} directly perceive the point cloud observations to produce a generalizable 3D scene-level representation for downstream control policy, effectively generalizing across different camera views through calibrations~\cite{rise}.
However, unlike the success of visual foundation models (VFMs)~\cite{moco, dino, clip} in images, the lack of generic pre-trained foundation models for 3D point clouds can limit the generalization capabilities of these methods at the object level, especially when facing objects with completely different geometries.

Alternatively, some literature~\cite{liv, vc1, r3m, mvp, theia} proposes decoupling perception and control in the policy by using pre-trained 2D visual representations for manipulations, but they offer limited advantages over VFMs in generalization evaluations~\cite{what_pretrain}.
SOFT~\cite{soft} investigates the use of intermediate features from VFMs and achieves competitive results against manipulation-specific pre-training, underscoring the capability of VFMs to enhance robot control policies in environmental perception with their extensive and diverse knowledge acquired from large-scale pre-training. 

To this end, we introduce \model, a data-efficient and generalizable robotic manipulation policy enabled by causal attention. With approximately 50 demonstrations from a single training environment, \model~can sustain its performance in test environments with unseen objects, backgrounds, and camera views. \model~integrates DINOv2~\cite{dino} with LoRA~\cite{lora} fine-tuning to perceive image observations, efficiently extracts manipulation-relevant features from raw observation tokens via the causal perceiver, and employs fine-grained attention-based conditioning to apply these features within the control policy. Extensive experiments demonstrate that \model~can effectively generalize across different levels of distribution shifts, and outperforms prior methods by a large margin.
\section{Related Works}

\subsection{Generalizable Robotic Manipulation}

Current generalist robotic manipulation policies~\cite{mt-act, rt1, oxe, openvla, peract, octo, gnfactor, same, rt2} are mainly trained under a multi-task setting using extensive robot demonstration data~\cite{mt-act, rt1, oxe, peract}, and generalization in these policies is predominantly discussed in the context of unseen tasks. Given the multi-task training setup and large volumes of data, it can be straightforward to find similar task combinations within the training data that may resemble the ``unseen'' tasks. As a result, as previously discussed in \S\ref{sec:intro}, it is challenging to determine whether the policy truly generalizes.

Therefore, we focus on a single-task setup and elevate the task to an abstract skill, making it adaptable to different target objects. While 2D image-based policies Diffusion Policy (DP)~\cite{dp} and ACT~\cite{act} perform well during in-domain evaluations, their generalization abilities are limited to minor environmental variations. In contrast, 3D point-cloud-based policies~\cite{rise, dp3d} can naturally handle spatial generalization across camera views, so long as the cameras are well-calibrated. Among these policies, RISE~\cite{rise} thoroughly evaluates its generalization ability in the real world and demonstrates state-of-the-art generalization performance across various environmental disturbances.

In this paper, we meticulously design experiments to assess the generalization performance of robotic manipulation policies. The highest level of generalization corresponds to testing in an entirely unseen environment, which rigorously evaluates the true generalization capability of the policy.

\subsection{Visual Representation Learning for Manipulation} 

Visual representation learning for manipulation primarily aims to generate useful and meaningful representations of current observation images, which are then used as inputs for control policies.

One line of methods decouples visual representation learning from the control policy. Some~\cite{liv, vc1, r3m, mvp} focus on pre-training vision encoders~\cite{vit, resnet} on large-scale human manipulation datasets~\cite{epickitchen, sth-sth, ego4d, doh100} using self-supervision schemes~\cite{byol, mae} for manipulation policies. However, recent studies~\cite{what_pretrain, soft} show that generic pre-trained VFMs~\cite{sam, dino, clip} can achieve comparable or even better generalization performance than these pre-trained visual representations.
Others investigate directly using VFM representations~\cite{what_pretrain, kat, soft} or distilling VFM representations~\cite{theia} for downstream learning, but it may overlook task-specific features in visually complex tasks without the fine-tuning process.
VINN~\cite{vinn} first trains visual representations from scratch using the demonstration dataset, and subsequently uses these representations to retrieve actions from the dataset, but its generalization capability is limited due to representation learning being restricted to the demonstration dataset domain.

Another line of research trains the visual encoder and the control policy together in an end-to-end manner. Some initialize pre-trained weights to the visual encoder for generalization considerations~\cite{umi, octo}. 
SpawnNet~\cite{spawnnet} designs adapter layers to integrate intermediate representations from both the encoder and a pre-trained encoder, enhancing generalization performance, but the large number of trainable parameters poses challenges in applying it to VFMs. 

Inspired by recent advances in parameter-efficient fine-tuning of foundation models, we apply LoRA~\cite{lora} for fine-tuning the VFM (DINOv2~\cite{dino}) to enhance observation perception effectively.
Perhaps the most similar to our work is SAM-E~\cite{same}, which also leverages the VFM (SAM~\cite{sam}) with LoRA fine-tuning for image perception. Nonetheless, as demonstrated in \S\ref{sec:ablation}, naively substituting visual encoders with VFMs yields limited generalization improvements, underscoring the effectiveness of our entire policy architecture. 

\section{Method}
\label{sec:model}

\begin{figure*}[t]
    \centering
    \includegraphics[width=0.85\textwidth]{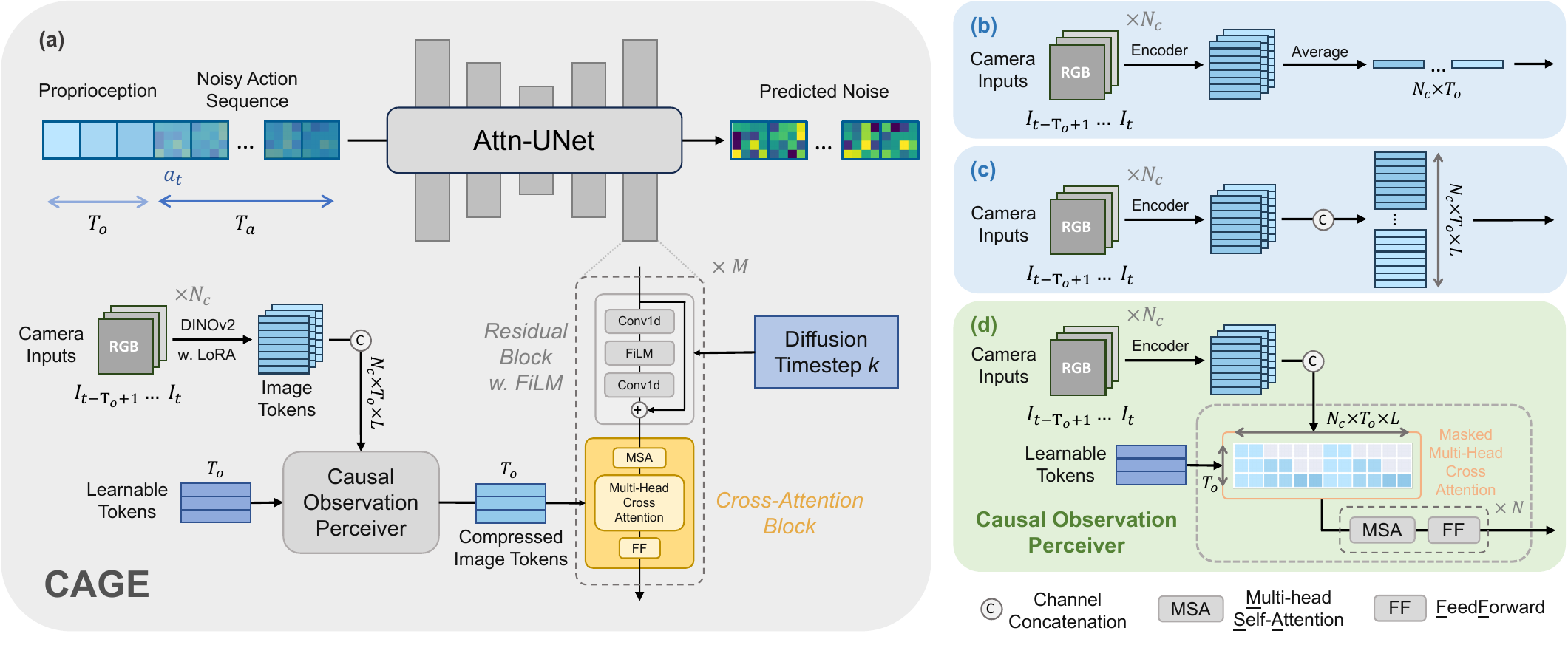}
    \caption{\textbf{(a) Overview of \model. } \model~is composed of three parts: 1) observation images are passed to DINOv2 image encoder (LoRA fine-tuning) to obtain observation tokens. 2) The concatenated observation tokens are compressed by \textit{Causal Observation Perceiver}. The learned tokens, along with the timestep embeddings serve as the conditioning for noise prediction. 3) The \textit{Attn-UNet} takes as input a sequence of noisy actions prefixed by proprioceptions and outputs noise estimation, following standard diffusion procedure. \textbf{(b) Averaging over tokens for compression} losses critical scene-level positional information. \textbf{(c) Directly using as downstream inputs} is inefficient due to the large number of tokens. \textbf{(d) Our proposed \textit{Causal Observation Perceiver} for token compression}.}
    \label{fig:model}
    \vspace{-0.4cm}
\end{figure*}

At current timestep $t$, with the observation horizon $T_o$, the robot receives stacks of images $I_t\in\mathbb{R}^{N_c\times T_o\times H\times W\times 3}$ from $N_c$ RGB cameras and proprioceptions $p_t\in\mathbb{R}^{T_o\times D_p}$ within the latest $T_o$ horizon, where $H\times W$ is the image size and $D_p$ is the proprioception dimension. Our \model~policy $f_\phi$ with parameters $\phi$ takes $I_t$ and $p_t$ as inputs and generate a continuous action sequence $a_t$ starting from $t$, with action horizon $T_a$, \textit{i.e.}, 
\begin{equation}
a_t = f_\phi(I_t, p_t)\in\mathbb{R}^{T_a\times D_a},
\end{equation}
where $D_a$ is the action dimension. Beyond learning from collected demonstrations, \textit{generalizing to similar and unseen environments} is also essential for robust robotic manipulation policies, especially in the real world.

Due to the promising performance of DP~\cite{dp} in robotic manipulations, we build our \model~upon it, incorporating 3 key design components for better generalization capability: DINOv2~\cite{dino} image encoder with LoRA~\cite{lora} fine-tuning (\S \ref{sec:dinov2}), causal perceiver for token compression (\S \ref{sec:perceiver}), and UNet diffusion action head with attention for effective conditioning (\S \ref{sec:action_head_attn}). Experiments show that all three parts are critical for the overall generalization ability. The architecture of \model~is shown in Fig.~\ref{fig:model}.

\subsection{Visual Foundation Model as Image Encoder}
\label{sec:dinov2}

Most prior studies~\cite{mt-act, rt1, dp, robomimic, act} prefer to utilize a lightweight CNN as the observation encoder for images, with these encoders typically being trained from scratch. This approach, while computationally efficient, has been found to result in rapid overfitting to the image domain of the training environment, restricting the generalization ability of the policy to even moderate visual variations that deviate from the training data. 
Recently, researchers have been investigating the use of ViT~\cite{vit} as image encoders to enhance policy performance~\cite{umi, maniwav}, but naively training ViT from scratch has shown limited improvement in the generalization capabilities of the policy.

In pursuit of a generalizable robotic manipulation policy, we utilize the VFM DINOv2~\cite{dino} (denoted as $h_\psi$ with parameters $\psi$) as our vision backbone due to its ability to extract rich and semantically meaningful features for downstream visual tasks. Unlike prior approaches that directly use DINOv2 features for downstream learning~\cite{what_pretrain, kat, soft}, we adopt a different strategy by keeping the pre-trained weights $\psi^*$ of DINOv2 frozen and employing LoRA~\cite{lora} for parameter-efficient fine-tuning with additional tuning parameters $\psi_\text{LoRA}$. This approach allows us to maintain the emergent segmentation capabilities of DINOv2 in out-of-distribution environments~\cite{what_pretrain, emerging}, while simultaneously adapting it for robotic manipulations to extract more task-relevant features. Specifically, our DINOv2 image encoder produces the following tokens
\begin{equation}
E_t = h_{\psi^* + \psi_\text{LoRA}} (I_t) \in \mathbb{R}^{(N_c\times T_o\times L) \times D_\text{token}},
\end{equation}
where $L$ is the number of tokens in a single input image, $D_\text{token}$ is the token dimension.

\subsection{Causal Perceiver for Token Compression}
\label{sec:perceiver}

\begin{figure*}
    \centering
    \includegraphics[width=0.82\linewidth]{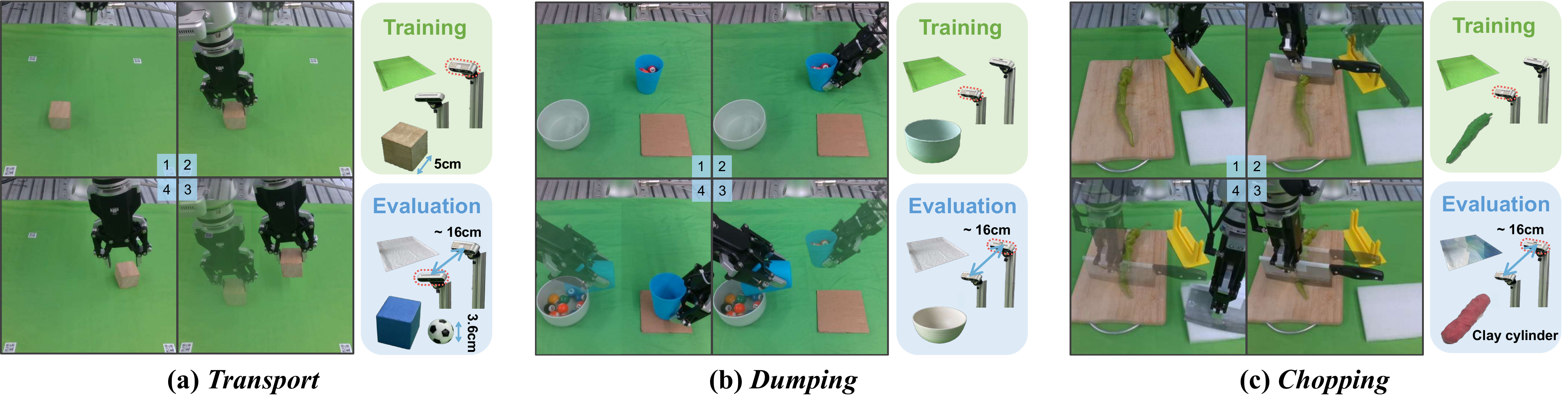}
    \vspace{-0.1cm}
    \caption{\textbf{Tasks with Generalization Variations.} \textbf{(a) \textit{Transport}}: the robot needs to localize the object on the left side of the workspace (\textit{reach}), grasp it (\textit{grasp}), and drop it to the right side (\textit{drop}). \textbf{(b) \textit{Dumping}}: the robot needs to grasp the cup horizontally (\textit{grasp}), move it over the bowl, tilt it until all balls are out into the bowl (\textit{pour}), and place the cup in the target area (\textit{place}). \textbf{(c) \textit{Chopping}}: the robot needs to grasp the kitchen knife (\textit{grasp}), use the knife to chop the object\protect\footnotemark on the cutting board into several pieces (\textit{chop}), and place the knife safely on the foam pad (\textit{place}). \textbf{Generalization variations} include different backgrounds, objects, and camera views (please refer to \S\ref{sec:gen-level} for detailed explanations).}
    \label{fig:task}
    \vspace{-0.4cm}
\end{figure*}

After processing the image observations via the encoder, we obtain $N_c\times T_o\times L$ image tokens in $E_t$. Common practices for handling such image tokens include averaging over them (Fig.~\ref{fig:model}(b))~\cite{what_pretrain, robomimic, soft}, or using the entire token sequence as input (Fig.~\ref{fig:model}(c))~\cite{mt-act, rt1, octo, act, rt2}. Yet, both methods present challenges in our context. Averaging loses critical scene-level positional information while using the entire token sequence is inefficient due to the large number of tokens. 

Inspired by \cite{peract, dexcap}, we propose a causal PerceiverIO-like~\cite{perceiver} structure (denoted as $g_\xi$ with parameters $\xi$) to efficiently compress the image tokens and extract the most relevant features for manipulation from the image tokens $E_t$. 
The $N_c\times T_o\times L$ image tokens from $N_c$ cameras within the observation horizon $T_o$ are first cross-attended by $T_o$ learnable tokens to compress into $T_o$ tokens within the observation horizon, with causal attention mask shown in Fig.~\ref{fig:model}(d). Then the output tokens are fed into $N$ causal transformer blocks to further process the extracted information. Formally, the causal perceiver $g_\xi$ compresses $N_c\times T_o\times L$ image tokens $E_t$ into $T_o$ tokens $E_t^*$, with
\begin{equation}
E_t^* = g_\xi(E_t) \in \mathbb{R}^{T_o\times D_\text{token}}.
\end{equation}

In this way, the compressed tokens $E_t^*$ can effectively retain both spatial and temporal information, which are equally important for manipulation tasks. 

\subsection{Diffusion Action Head with Attention}
\label{sec:action_head_attn}

The original UNet-based~\cite{unet} diffusion action head~\cite{dp} employs FiLM~\cite{film} conditioning mechanics for the observation inputs. However, the FiLM module is non-expandable as it only accepts 1D input, limiting the expressivity of the observation conditions. Simply flattening the compressed tokens into one dimension will ignore the temporal relationship among $T_o$ compressed tokens in the observation horizon.

To mitigate this issue, we propose to decouple the conditioning of diffusion step $k$ and compressed image tokens $E_t^*$. By employing a structure that incorporates a residual block~\cite{resnet} followed by a cross-attention block~\cite{transformer}, which is similar to the 2D UNet backbone in Stable Diffusion~\cite{sd}, we can apply diffusion step $k$ conditioning in the residual block via FiLM~\cite{film} and use cross-attention to condition on the compressed image tokens $E_t^*$. FiLM offers efficient global conditioning with the diffusion step, while cross-attention enables fine-grained conditioning with image tokens.

The UNet-based diffusion action head (denoted as $\epsilon_\varphi$ with parameters $\varphi)$ is modified accordingly, as illustrated in Fig.~\ref{fig:model}, which we refer to as \textit{Attn-UNet}. The action is then iteratively denoised using the DDIM~\cite{ddim} schedule.
\begin{equation}
a_t^{(k-1)}=\alpha_ka_t^{(k)}-\beta_k\epsilon_\varphi\left(a_t^{(k)}; E_t^*, k\right)  \in \mathbb{R}^{T_a\times D_a},
\end{equation}
where $k$ is the diffusion step, $\{\alpha_k,\beta_k\}$ is the noise schedule, and $a_t=a_t^{(0)}$ is the generated continuous action trajectory.

In summary, \model~is parameterized by the combination of the trainable parameters in each module, with a frozen DINOv2 weights $\psi^*$, \textit{i.e.}, $\phi = (\psi_\text{LoRA},\xi,\varphi; \psi^*)$.

\section{Experiments}

\begin{figure*}
\begin{adjustbox}{valign=t,minipage={0.58\linewidth}}
    \includegraphics[height=0.2276\linewidth]{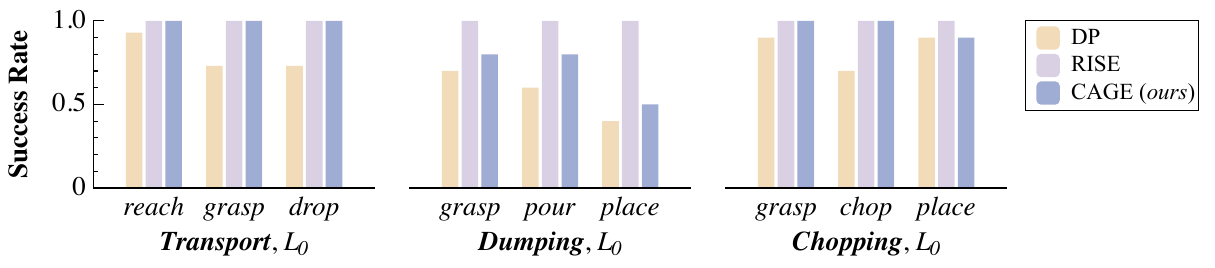}
\end{adjustbox}
\hfill
\begin{adjustbox}{valign=t,minipage={0.42\linewidth}}
    \centering
    \footnotesize
    \setlength\tabcolsep{3pt}
    \begin{tabular}{cccc}
        \toprule
        \multirow{2}{*}{\textbf{Method}~/~\textbf{Task}} & \multicolumn{2}{c}{\textbf{\textit{Dumping}}} & \textbf{\textit{Chopping}} \\
        \cmidrule(lr){2-3} \cmidrule(lr){4-4}
        &\# ball & \# ball (g) & \# chop \\
        \midrule
        DP~\cite{dp} & 4.2\,/\,10 & 7.0\,/\,10 & 2.0\,/\,4 \\
        RISE~\cite{rise} & \best{9.3}\,/\,10 & 9.3\,/\,10 & \best{2.8}\,/\,4 \\
        \midrule
        \model~(\textit{ours}) &  7.5\,/\,10 & \best{9.4}\,/\,10 & 2.3\,/\,4 \\
        \bottomrule
    \end{tabular}%
\end{adjustbox}
\captionof{table}{\textbf{Results of the $L_0$ Evaluation.} The $L_0$ evaluation is conducted in the same environments seen during training. \model~outperforms DP in all tasks, and performs comparably with RISE in most tasks.}
\label{tab:base}
\vspace{-0.1cm}
\end{figure*}

\begin{figure*}
\begin{adjustbox}{valign=c,minipage={0.58\linewidth}}
    \includegraphics[height=0.2276\linewidth]{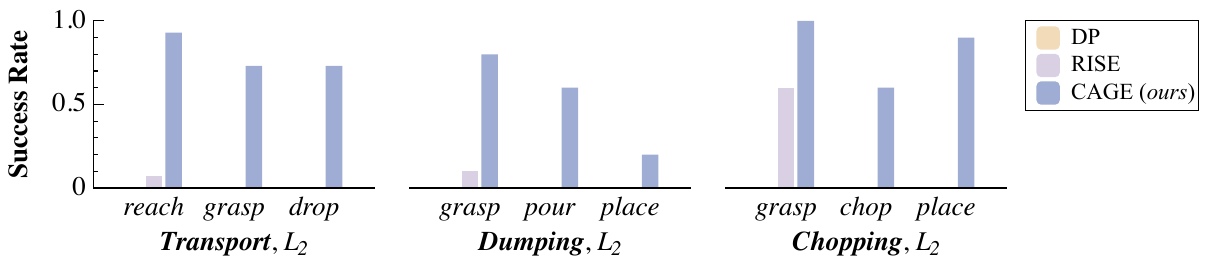}
\end{adjustbox}
\hfill
\begin{adjustbox}{valign=c,minipage={0.42\linewidth}}
    \centering
    \footnotesize
    \setlength\tabcolsep{3pt}
    \begin{tabular}{cccc}
        \toprule
        \multirow{2}{*}{\textbf{Method}~/~\textbf{Task}} & \multicolumn{2}{c}{\textbf{\textit{Dumping}}} & \textbf{\textit{Chopping}} \\
        \cmidrule(lr){2-3} \cmidrule(lr){4-4}
        &\# ball & \# ball (g) & \# chop \\
        \midrule
        DP~\cite{dp} & 0.0\,/\,10 & 0.0\,/\,10 & 0.0\,/\,4 \\
        RISE~\cite{rise} & 0.0\,/\,10 & 0.0\,/\,10 & 0.0\,/\,4 \\
        \midrule
        \model~(\textit{ours}) &  \best{3.4}\,/\,10 & \best{5.7}\,/\,10 & \best{0.9}\,/\,4 \\
        \bottomrule
    \end{tabular}%
\end{adjustbox}
\captionof{table}{\textbf{Results of the $L_2$ Evaluation.} The $L_2$ evaluation is conducted in unseen environments, with different backgrounds, camera views and unseen objects. \model~largely outperforms all baselines in all tasks.}
\label{tab:l2}
\vspace{-0.1cm}
\end{figure*}

\begin{figure*}
\begin{adjustbox}{valign=t,minipage={\linewidth}}
    \includegraphics[height=0.132\linewidth]{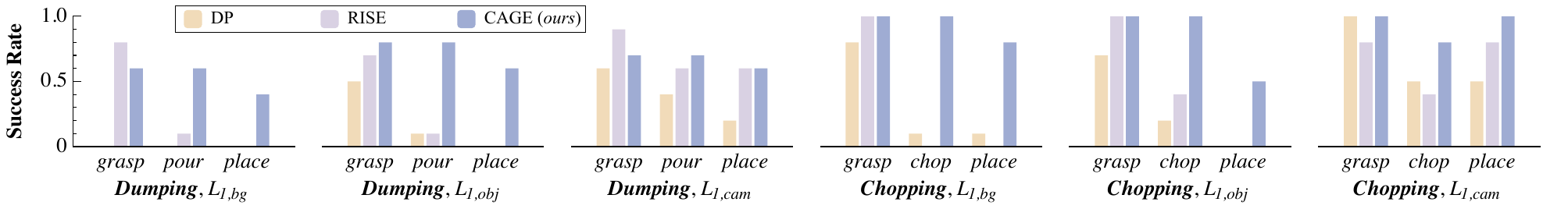}
    \vspace{0.15cm}
\end{adjustbox}
\begin{adjustbox}{valign=t,minipage={\linewidth}}
    \centering  
    \footnotesize
    \setlength\tabcolsep{4.5pt}
    \begin{tabular}{ccccccccccccc}
        \toprule
        \multirow{3}{*}{\textbf{Method}} & \multicolumn{3}{c}{\textbf{\textit{Transport}}} & \multicolumn{6}{c}{\textbf{\textit{Dumping}}} & \multicolumn{3}{c}{\textbf{\textit{Chopping}}}  \\
        \cmidrule(lr){2-4}\cmidrule(lr){5-10}\cmidrule(lr){11-13}
        & \multirow{2}{*}{$L_{1,bg}$} & \multirow{2}{*}{\begin{tabular}{c}$L_{1,obj}$\\ (block / ball)\end{tabular}} & \multirow{2}{*}{$L_{1,cam}$} & \multicolumn{2}{c}{$L_{1,bg}$} & \multicolumn{2}{c}{$L_{1,obj}$} & \multicolumn{2}{c}{$L_{1,cam}$} & $L_{1,bg}$ & $L_{1,obj}$ & $L_{1,cam}$ \\
        \cmidrule(lr){5-6}\cmidrule(lr){7-8}\cmidrule(lr){9-10}\cmidrule(lr){11-11}\cmidrule(lr){12-12}\cmidrule(lr){13-13}
            & & & & \# ball & \# ball (g) & \# ball & \# ball (g) & \# ball & \# ball (g) & \# chop & \# chop & \# chop \\
        \midrule
        DP~\cite{dp} & 0.00 & 0.00 / 0.00 & 0.27 & 0.0\,/\,10 & 0.0\,/\,10 & 0.2\,/\,10 & 0.4\,/\,10 & 1.9\,/\,10 & 3.2\,/\,10 & 0.1\,/\,4 & 0.3\,/\,4 & 1.2\,/\,4 \\
        RISE~\cite{rise} & 0.13 & 0.00 / 0.00 & 0.60 & 0.8\,/\,10 & 1.0\,/\,10 & 0.2\,/\,10 & 0.3\,/\,10 & 4.6\,/\,10 & 5.1\,/\,10 & 0.0\,/\,4 & 0.4\,/\,4 & 0.6\,/\,4 \\ \midrule
        \model~\textit{(ours)} & \textbf{0.87} & \textbf{0.80} / \textbf{0.67} & \textbf{0.80} & \textbf{4.4}\,/\,10 & \textbf{7.3}\,/\,10 & \textbf{6.2}\,/\,10 & \textbf{7.8}\,/\,10 & \textbf{5.7}\,/\,10 & \textbf{8.1}\,/\,10 & \textbf{2.2}\,/\,4 & \textbf{1.8}\,/\,4 & \textbf{1.5}\,/\,4 \\
        \bottomrule
    \end{tabular}
\end{adjustbox}
\captionof{table}{\textbf{Results of the $L_1$ Evaluation.} The $L_1$ evaluation is conducted in similar environments, with one of the variations in backgrounds, objects and camera views. The overall success rates are reported for the \textbf{\textit{Transport}} task here. \model~surpasses RISE on success rates and completion rates across all tasks.}
\label{tab:l1}
\vspace{-0.8cm}
\end{figure*}

\begin{table*}
\end{table*}

\subsection{Setup}

\subsubsection{Platform}
\label{sec:platform}

We employ a Flexiv Rizon robotic arm~\cite{flexiv} with a Dahuan AG-95 gripper~\cite{dahuan} as our robot platform. A rectangle area of $45\text{cm} \times 60\text{cm}$ in the center of the tabletop is selected as the robot workspace. We adopt a binocular vision setting with 2 Intel RealSense D435 cameras~\cite{realsense}: one fixed camera in front of the robot facing towards the workspace, and one in-hand camera above the gripper on the end of the robotic arm. 

\footnotetext{We use real peppers during data collection. However, to avoid unnecessary waste, we make a pepper out of clay for evaluation.}

\subsubsection{Baselines}

We choose the 2D image-based Diffusion Policy (DP)~\cite{dp}, and the current state-of-the-art 3D point-cloud-based policy RISE~\cite{rise} with decent generalization performance in the real world as baselines. For DP, RGB images from both cameras are used, whereas the RGB-D frames from only the fixed camera are provided to RISE following its original implementation. Both methods are trained from scratch with their recommended hyper-parameter settings. Please refer to Appendix~\ref{appendix:impl} for details.

\subsubsection{Tasks}

In order to provide a comprehensive coverage of robotic manipulation problems, we design 3 tasks with \textit{generalizable skills} (illustrated in Fig.~\ref{fig:task}), each focusing on a different aspect of robotic manipulation: \textbf{\textit{Transport}} (Pick-and-Place), \textbf{\textit{Dumping}} (6-Dof) and \textbf{\textit{Chopping}} (Long horizon). With a special focus on generalization, our tasks are designed at the skill level, so we do not impose strict requirements on the objects, and they can be substituted with any similar items during evaluations.

\subsubsection{Demonstrations}

Following~\cite{rh20t}, human-teleoperated demonstrations are collected via a haptic device~\cite{sigma7}. We collect 50 demonstrations for the \textit{\textbf{Transport}} task, and 40 demonstrations for the \textit{\textbf{Dumping}} and the \textit{\textbf{Chopping}} tasks each. 
To properly assess the real generalization capabilities of each policy, the variations of background, object, and camera view are deliberately limited in demonstrations for a more restricted training environment.

\subsubsection{Evaluation Protocols}

Evaluations are conducted on a workstation with an Intel Core i9-10900K CPU and an NVIDIA RTX 3090 GPU linked to the robot platform. During the evaluation, we run the inference process and the execution process in parallel for continuous control with a frequency of 10Hz. During the evaluations, we randomly and uniformly generate test positions beforehand for each task and initialize the workspace accordingly, ensuring a fair, consistent, and reproducible comparison across different methods. We report stage-wise success rates (the stages are defined in the caption of Fig.~\ref{fig:task}) to evaluate the performance throughout the task. For the \textbf{\textit{Dumping}} and the \textbf{\textit{Chopping}} tasks, we also report the task completion rates, including the average number of balls poured into the bowl (\# ball), the average number of poured balls if successfully grasped (\# ball (g)), and the average number of chops (\# chop) respectively.

\subsubsection{Generalization Levels}
\label{sec:gen-level}
We meticulously devised 3 levels of tests to evaluate thoroughly each policy's generalization capabilities:
\begin{itemize}
    \item \textbf{\textit{L}$_\mathbf{0}$ generalization} refers to evaluations in the \textit{training environment} with the same background, seen objects, and the same camera view;
    \item \textbf{\textit{L}$_\mathbf{1}$ generalization} refers to evaluations in a \textit{similar environment} with the variation in one of the three aforementioned aspects, denoted as $L_{1, bg}$, $L_{1, obj}$ and $L_{1, cam}$ respectively;
    \item \textbf{\textit{L}$_\mathbf{2}$ generalization} refers to evaluations in an \textit{unseen environment} with variations present across all three aspects, thereby emulating the scenario in real-world deployment.
\end{itemize}

\subsubsection{Implementations}
\label{sec:impl}

For each task, \model~is trained with a batch size of 64 on 4 NVIDIA A100 80GB GPUs for 500 epochs. We set $N_c = 2$ based on the binocular vision settings in our robot platform. For the vision encoder, we choose \texttt{DINOv2-large}~\cite{dino} as the backbone and set the LoRA~\cite{lora} rank to 16. We set the observation horizon $T_o=4$ and the action horizon $T_a=20$ with a relative action representation~\cite{umi}.
Unlike prior works~\cite{dp, umi, act}, the data preprocessings are limited to color jitter and center crop to ensure a out-of-distribution evaluation of camera viewpoint generalization. However, we conduct experiments with additional geometric augmentations to show the full potential of \model~in \S\ref{sec:ablation}.

\subsection{Results}

\begin{table*}
\centering
\begin{tabular}{cccccccccc}
    \toprule
    \multicolumn{2}{c}{\textbf{Platform}} & \multicolumn{3}{c}{\textbf{\textit{Transport}}, \textit{similar}} & \multicolumn{5}{c}{\textbf{\textit{Dumping}}, \textit{similar}} \\
    \cmidrule(lr){1-2}\cmidrule(lr){3-5}\cmidrule(lr){6-10}
    \textbf{Robot} & \textbf{Gripper} & reach & grasp & drop & grasp & pour & place & \# ball & \# ball (g) \\
    \midrule
    Flexiv & Dahuan & 1.00 & 1.00 & 1.00 & 0.80 & 0.80 & 0.50 & 7.5\,/\,10 & 9.4\,/\,10 \\
    Flexiv & \cellcolor{yellow}Robotiq & 0.73 & 0.60 & 0.60 & 0.40 & 0.40 & 0.40 & 3.2\,/\,10 & 8.0\,/\,10 \\
    \cellcolor{yellow}RealMan & Dahuan & 0.53 & 0.47 & 0.47 & 0.20 & 0.20 & 0.10 & 1.2\,/\,10 & 6.0\,/\,10\\
    \bottomrule
\end{tabular}
\caption{\textbf{Cross-Embodiment Evaluation Results under \textit{Similar} Environment.} \textit{Similar} denotes a similar environment setup with a green background, the same objects, and a close camera view. The background and the camera are not identical to the training environment as their platforms are different. New hardware is highlighted in yellow.}
\label{tab:cross_1}
\vspace{-0.1cm}
\end{table*}

\begin{table*}
\centering
\begin{tabular}{cccccccccc}
    \toprule
    \multicolumn{2}{c}{\textbf{Platform}} & \multicolumn{3}{c}{\textbf{\textit{Transport}}, \textit{novel}} & \multicolumn{5}{c}{\textbf{\textit{Dumping}}, \textit{novel}} \\
    \cmidrule(lr){1-2}\cmidrule(lr){3-5}\cmidrule(lr){6-10}
    \textbf{Robot} & \textbf{Gripper} & reach & grasp & drop & grasp & pour & place & \# ball & \# ball (g) \\
    \midrule
    Flexiv & Dahuan & 0.93 & 0.73 & 0.73 & 0.80 & 0.60 & 0.20 & 3.4\,/\,10 & 4.3\,/\,10 \\
    Flexiv & \cellcolor{yellow}Robotiq & 0.60 & 0.53 & 0.53 & 0.40 & 0.40 & 0.40 & 3.2\,/\,10 & 8.0\,/\,10 \\
    \cellcolor{yellow}RealMan & Dahuan & 0.33 & 0.13 & 0.13 & 0.10 & 0.00 & 0.00 & 0.0\,/\,10 & 0.0\,/\,10\\
    \bottomrule
\end{tabular}
\caption{\textbf{Cross-Embodiment Evaluation Results under \textit{Novel} Environment.} We further replace the background and the target objects following the same protocol as described in \S\ref{sec:gen-level}. New hardware is highlighted in yellow.}
\label{tab:cross_2}
\vspace{-0.4cm}
\end{table*}

\subsubsection{Training Environment}

For the $L_0$ evaluations in the same environment, we present the results in Tab.~\ref{tab:base}.

\textbf{\model~consistently outperforms DP and performs comparably to RISE in most tasks under in-domain environments.} In the training environment, \model~surpasses DP in all tasks. RISE, which utilizes 3D point cloud information, generally performs the best among all methods. However, \model~still shows comparable performance to RISE in the \textit{\textbf{Transport}} and the \textit{\textbf{Chopping}} tasks. For the \textit{\textbf{Dumping}} task, precise cup localization is critical to avoid task failure due to collisions, and 2D image-based policy \model~are naturally less accurate in position prediction than 3D point-cloud-based policy RISE. But if we only consider the average number of balls poured over successful grasps, \model~actually performs the pouring action slightly better than RISE.

\subsubsection{Similar Environment}
\label{sec:similar}
For the visually altered $L_1$ evaluations, we design several challenging environments such as replacing the block with a ball of drastic differences in both size and geometry for the \textbf{\textit{Transport}} task. The results are shown in Tab.~\ref{tab:l1}.

\textbf{\model~demonstrates effective generalization across various similar environments, with an acceptable performance decline in general.} While subtle visual changes in the evaluation environments challenge both DP and RISE, preventing them from maintaining their in-domain performance, \model~consistently outperforms these baseline methods. Notably, in the \textbf{\textit{Transport}} task with a football as the target, \model~achieves an average success rate of 0.67 without any fine-tuning for this object, showcasing its potential to transfer manipulation skills to visually distinct objects.

\subsubsection{Unseen Environment} 

For the $L_2$ evaluations, we introduce distribution shifts on background, object, and camera view at the same time, resulting in the most challenging test environments. The results are illustrated in Tab.~\ref{tab:l2}.

\textbf{\model~exhibits a strong generalization ability even to unseen environments, efficiently completing the task in out-of-domain situations without the need for fine-tuning.}
\model~remains an average of over 0.5 success rate while other policies still struggle to recognize the target object in the environment. This further demonstrates the effectiveness of our approach in learning manipulation-specific tasks while retaining most of the object-centric knowledge within the pre-trained visual encoders.

\subsection{Cross-Embodiment Evaluations}
\label{sec:cross-embodiment}

Apart from visual generalization capabilities, it is essential for a generalizable policy to effectively perform the learned tasks on new embodiments. Despite the inherent challenges, we report the task success rates for the \textbf{\textit{Transport}} and \textbf{\textit{Dumping}} tasks, as well as the completion rates for the latter, on new platforms with different hardware configurations.

The evaluation results on two similar platforms featuring either a new gripper or a different type of robotic arm are presented in Tab.~\ref{tab:cross_1} and Tab.~\ref{tab:cross_2}. A significant performance drop is expected given the entirely new workspaces and manipulation interfaces. Nonetheless, for straightforward pick-and-place tasks like \textbf{\textit{Transport}}, \model~achieves a near 50\% success rate in most cases, except for configurations involving the RealMan arm in a \textit{novel} environment.

Additionally, we observe that for more delicate tasks, our model demonstrates greater robustness to changes in the gripper compared to variations in the robotic arm. While these results do not yet meet the standard of excellence, we hope they can serve as a valuable baseline for future efforts aimed at enhancing cross-embodiment generalization performance.

\subsection{RH20T Pre-training}

\begin{table}[t]
\hspace{0.1cm}
\begin{adjustbox}{valign=c,minipage={0.22\linewidth}}
    \vspace{0.1cm}
    \centering
    \includegraphics[width=\linewidth]{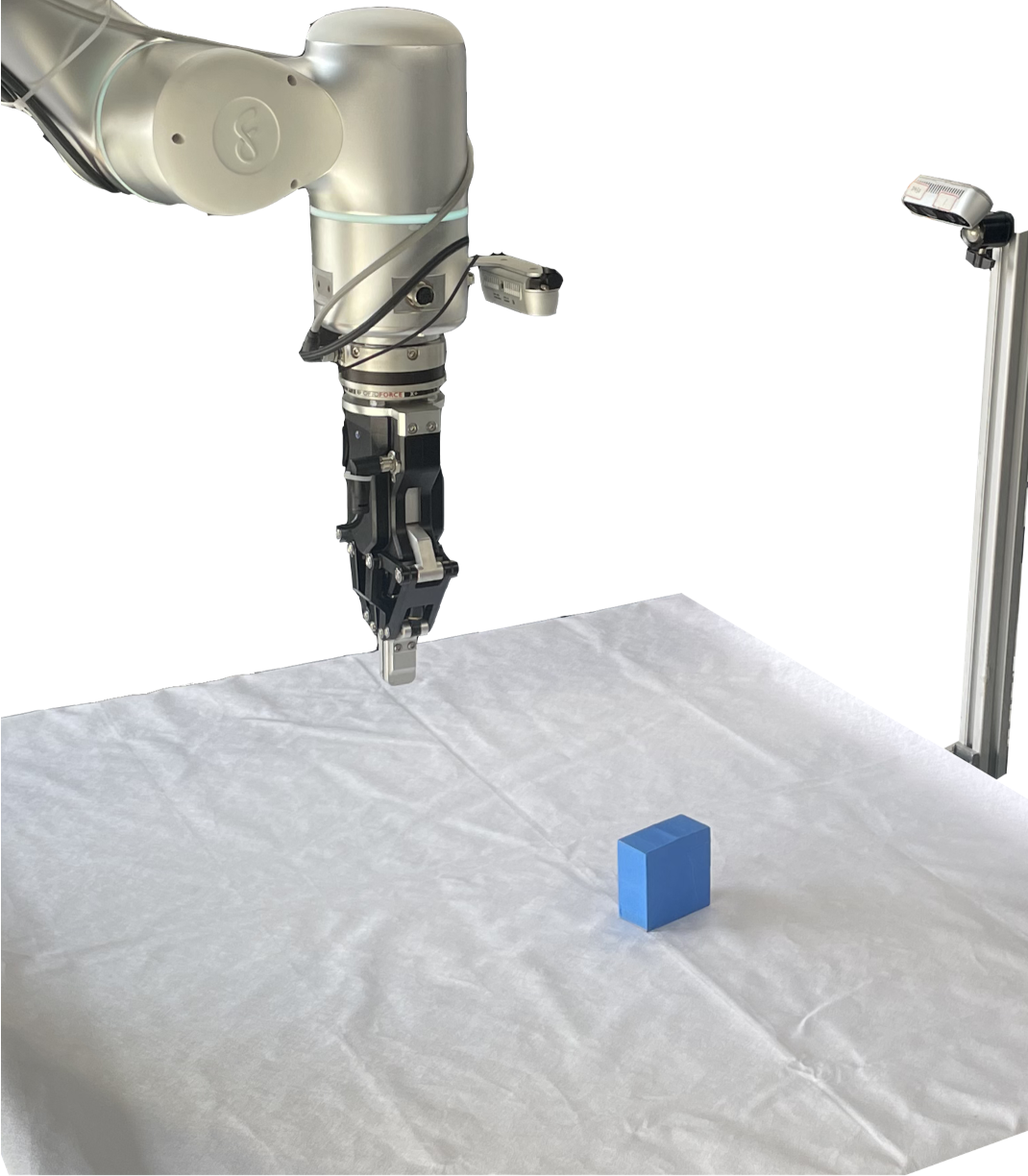}
\end{adjustbox}
\hfill
\begin{adjustbox}{valign=c,minipage={0.75\linewidth}}
    \footnotesize
    \setlength\tabcolsep{1pt}
    \centering
    \begin{tabular}{ccc}
        \toprule
        \begin{tabular}{c}\textbf{Success}\\ \textbf{Rate}\end{tabular} &  \begin{tabular}{c}\model\\ (\textit{semi-in-domain}) \end{tabular} & \begin{tabular}{c}\model\\(RH20T, \textit{out-of-box})\end{tabular} \\
        \midrule
        Reach & 0.93 & 0.80 ({\color{darkgreen}-0.13}) \\
        Grasp & 0.80 & 0.67 ({\color{darkgreen}-0.13}) \\
        Drop & 0.73 & 0.67  ({\color{darkgreen}-0.06})\\
        \bottomrule
    \end{tabular}
\end{adjustbox}
\caption{\textbf{Results of the Out-of-Box Evaluation.} The figure on the left illustrates the out-of-box environments for \model~pre-trained on RH20T~\cite{rh20t}. Note that none of the object, background, and camera views appear in the training dataset.}
\label{tab:rh100t}
\vspace{-0.5cm}
\end{table}

In addition to the self-collected, high-quality dataset, we train \model~on the same \textit{\textbf{Transport}} task from RH20T~\cite{rh20t} to evaluate the out-of-box generalization ability after pre-training. This dataset contains 164 demonstrations, collected by 14 operators under 4 different embodiments, and incorporates diverse distractors and rich variations in object sizes, backgrounds, and camera views. For evaluation, we design a blue cuboid\footnotemark~that differs from the training blocks both in texture and shape as the target. As illustrated in Tab.~\ref{tab:rh100t}, we observe that while baselines fail to perform the task, \textbf{\model~is capable of operating out-of-box and achieving comparable performance against being trained with a \textit{semi-in-domain} high-quality dataset} (only same in the camera view). This proves that our approach can be scaled up to large and diverse datasets and yield a corresponding performance improvement.

\footnotetext{The originally purposed test object (football) is already used as a distractor in this dataset.}

\subsection{Ablations}
\label{sec:ablation}

\begin{table*}
\centering
\begin{tabular}{ccccccccccccc}
\toprule
\multirow{2}{*}{\textbf{Ablation}} &
  \multicolumn{3}{c}{$L_0$} &
  \multicolumn{3}{c}{$L_{1,bg}$} &
  \multicolumn{3}{c}{$L_{1,obj}$} &
  \multicolumn{3}{c}{$L_{1,cam}$} \\
  \cmidrule(lr){2-4}\cmidrule(lr){5-7}\cmidrule(lr){8-10}\cmidrule(lr){11-13}
                               & reach & grasp & drop & reach & grasp & drop & reach & grasp & drop & reach & grasp & drop \\
                               \midrule
\model~(\textit{ours}) & 1.00  & 1.00    & 1.00 & 0.87  & 0.87    & 0.87 & 0.93  & 0.80    & 0.80 & 1.00  & 0.80    & 0.80 \\
\midrule
- \textit{Attn-UNet} & 0.80  & 0.80    & 0.80 & 0.80  & 0.67    & 0.67 & 0.73  & 0.73    & 0.73 & 0.80  & 0.67    & 0.67 \\
- Causal Obs. Perceiver & 0.47  & 0.40    & 0.40 & 0.40  & 0.13    & 0.13 & 0.27  & 0.00    & 0.00 & 0.40  & 0.20    & 0.20 \\
- DINOv2\protect\footnotemark & 0.93  & 0.73    & 0.73 & 0.00  & 0.00    & 0.00 & 0.00  & 0.00    & 0.00 & 0.67  & 0.53    & 0.47 \\
\midrule
DP~\cite{dp} & 0.93  & 0.73    & 0.73 & 0.00  & 0.00    & 0.00 & 0.00  & 0.00    & 0.00 & 0.53  & 0.27    & 0.27 \\ \bottomrule
\end{tabular}
\caption{\textbf{Ablation Results in \textit{Transport} Task.} Three key components are sequentially removed from \model.}
\label{tab:ablation}
\vspace{-0.2cm}
\end{table*}

\subsubsection{Model Components}

As shown in Tab.~\ref{tab:ablation}, we observe significant performance gains across the base and $L_1$ levels of evaluations with the introduction of Causal Observation Perceiver (2nd row \textit{v.s.} 3rd row). This validates our statements in \S\ref{sec:model} that \textbf{Causal Observation Perceiver can effectively retain both spatial and temporal information, facilitating the extraction of features for manipulation}. 

Interestingly, replacing the original ResNet~\cite{resnet} image encoder to DINOv2~\cite{dino} hinders the performance, supporting the claim that end-to-end training yields better in-domain performance~\cite{dp, octo}. However, the model with DINOv2 shows improved ability to recognize targets with texture changes ($L_{1, bg}$ and $L_{1, obj}$), resulting in a higher success rate for reach (3rd row \textit{v.s.} 4th row).

Moreover, the addition of attention modules also brings considerable improvements in all tests, with an average of more than 20\% increase. This proves that \textbf{decoupling diffusion timestep with observation inputs and keeping the temporal relationship of history observations can help the policy learn generalizable manipulation skills}.


\subsubsection{Geometric Augmentation}

\begin{table}[t]
    \centering
    \setlength\tabcolsep{3pt}
    \begin{tabular}{cccc}
        \toprule
        \textbf{\textit{w/wo}. Geo. Aug.} & $L_{1,\text{cam}}$ (upper left) & $L_{1,\text{cam}}$ (lower right) & $L_2$ \\
        \midrule
         &  0.80 & 0.80 & 0.40 \\
        \checkmark & \best{0.87} & \best{0.93} & \best{0.73} \\
        \bottomrule
    \end{tabular}
    \caption{\textbf{Ablation Results \textit{w/wo} Geometric Augmentations}.}
    \label{tab:geo_aug}
    \vspace{-0.5cm}
\end{table}

In our initial setup (\S\ref{sec:impl}), we intentionally exclude random cropping and other augmentation methods, creating a mono-distributed training environment. We now unlock the full potential of \model~by applying geometric augmentations, including random perspective transform followed by random cropping. The results shown in Tab.~\ref{tab:geo_aug} demonstrate a significant boost in generalization abilities, notably with an 82.5\% increase in success rate in the $L_2$ evaluation. Therefore, for better performance in the challenging $L_2$ evaluations, we include these geometric augmentations for all image-based policies.

\subsubsection{Observation Horizon}

Unlike DP~\cite{dp}, \model~can handle long observation horizon from multiple cameras efficiently. We show the performance comparisons of the observation horizon $T_o$ in Tab.~\ref{tab:Ho}, and find that $T_o=4$ performs better than the original DP implementation ($T_o=2$), demonstrating the superiority of our approach in effectively extracting and utilizing temporal information from a larger history window.

\footnotetext{After removing DINOv2, the remaining model is still not identical to DP~\cite{dp}, as we opt for an observation horizon of 4 rather than 2 and use proprioception as the prefix to the noisy actions. Please refer to Appendix~\ref{appendix:impl} for detailed explanations.}

\section{Conclusion}

In this work, we introduce \model, a data-efficient robotic manipulation policy that demonstrates strong generalization abilities in handling different levels of distribution shifts. We show that the success of \model~is based on the collaboration of three key components: The vision foundation model as encoder, a causal observation perceiver, and an attention-augmented UNet as the action predictor. Through extensive experiments, \model~not only outperforms DP~\cite{dp}, but also surpasses 3D-based RISE~\cite{rise} by a considerable margin in terms of generalization. 
Despite \model's success, we have observed several possible directions for future improvements. 
For example, in out-of-box evaluations, we find that \model~performs better within the workspace of RH20T (7/8) than those positions outside (3/7), suggesting that data augmentations at the trajectory level might be needed for an enhanced workspace generalizability.
Moreover, on the \textbf{\textit{Chopping}} task, \model~as well as baselines all fail to chop cleanly, resulting in connected pieces. After examination, we find that such difficulty also exists for human operators in the absence of force feedback from the haptic device. This highlights the necessity for the control policy to incorporate real-time force data for such contact-rich tasks.

\section*{Acknowledgement}

\begin{table}[t]
    \centering
    \begin{tabular}{cccc}
        \toprule
        \multirow{2}{*}{\begin{tabular}{c}\textbf{Observation}\\ \textbf{Horizon $T_o$}\end{tabular}} &
          \multicolumn{3}{c}{\textbf{\textit{Transport}}, $L_0$}  \\
          \cmidrule(lr){2-4}
         & reach & grasp & drop \\
        \midrule
        2 &  \best{1.00} & 0.87 & 0.87 \\
        4 & \best{1.00} & \best{1.00} & \best{1.00} \\
        8 & 0.87 & 0.87 & 0.87 \\
        \bottomrule
    \end{tabular}
    \caption{\textbf{Ablation Results of the Observation Horizon.}}
    \label{tab:Ho}
    \vspace{-0.5cm}
\end{table}

We would like to thank Chenxi Wang for insightful discussions and for setting up the RealMan platform in cross-embodiment evaluations, Yiming Wang and Zihao He for their help during the data collection process.

\printbibliography

@inproceedings{dp,
  author       = {Cheng Chi and
                  Siyuan Feng and
                  Yilun Du and
                  Zhenjia Xu and
                  Eric Cousineau and
                  Benjamin Burchfiel and
                  Shuran Song},
  editor       = {Kostas E. Bekris and
                  Kris Hauser and
                  Sylvia L. Herbert and
                  Jingjin Yu},
  title        = {Diffusion Policy: Visuomotor Policy Learning via Action Diffusion},
  booktitle    = {Robotics: Science and Systems XIX, Daegu, Republic of Korea, July
                  10-14, 2023},
  year         = {2023},
}

@inproceedings{ddim,
  author       = {Jiaming Song and
                  Chenlin Meng and
                  Stefano Ermon},
  title        = {Denoising Diffusion Implicit Models},
  booktitle    = {9th International Conference on Learning Representations, {ICLR} 2021,
                  Virtual Event, Austria, May 3-7, 2021},
  publisher    = {OpenReview.net},
  year         = {2021},
}

@inproceedings{act,
  author       = {Tony Z. Zhao and
                  Vikash Kumar and
                  Sergey Levine and
                  Chelsea Finn},
  editor       = {Kostas E. Bekris and
                  Kris Hauser and
                  Sylvia L. Herbert and
                  Jingjin Yu},
  title        = {Learning Fine-Grained Bimanual Manipulation with Low-Cost Hardware},
  booktitle    = {Robotics: Science and Systems XIX, Daegu, Republic of Korea, July
                  10-14, 2023},
  year         = {2023},
}

@article{openvla,
  title={OpenVLA: An Open-Source Vision-Language-Action Model},
  author={Kim, Moo Jin and Pertsch, Karl and Karamcheti, Siddharth and Xiao, Ted and Balakrishna, Ashwin and Nair, Suraj and Rafailov, Rafael and Foster, Ethan and Lam, Grace and Sanketi, Pannag and others},
  journal={arXiv preprint arXiv:2406.09246},
  year={2024}
}

@inproceedings{r3m,
  author       = {Suraj Nair and
                  Aravind Rajeswaran and
                  Vikash Kumar and
                  Chelsea Finn and
                  Abhinav Gupta},
  editor       = {Karen Liu and
                  Dana Kulic and
                  Jeffrey Ichnowski},
  title        = {{R3M:} {A} Universal Visual Representation for Robot Manipulation},
  booktitle    = {Conference on Robot Learning, CoRL 2022, 14-18 December 2022, Auckland,
                  New Zealand},
  series       = {Proceedings of Machine Learning Research},
  volume       = {205},
  pages        = {892--909},
  publisher    = {{PMLR}},
  year         = {2022},
}

@inproceedings{liv,
  author       = {Yecheng Jason Ma and
                  Vikash Kumar and
                  Amy Zhang and
                  Osbert Bastani and
                  Dinesh Jayaraman},
  editor       = {Andreas Krause and
                  Emma Brunskill and
                  Kyunghyun Cho and
                  Barbara Engelhardt and
                  Sivan Sabato and
                  Jonathan Scarlett},
  title        = {{LIV:} Language-Image Representations and Rewards for Robotic Control},
  booktitle    = {International Conference on Machine Learning, {ICML} 2023, 23-29 July
                  2023, Honolulu, Hawaii, {USA}},
  series       = {Proceedings of Machine Learning Research},
  volume       = {202},
  pages        = {23301--23320},
  publisher    = {{PMLR}},
  year         = {2023},
}

@article{dino,
  author       = {Maxime Oquab and
                  Timoth{\'{e}}e Darcet and
                  Th{\'{e}}o Moutakanni and
                  Huy V. Vo and
                  Marc Szafraniec and
                  Vasil Khalidov and
                  Pierre Fernandez and
                  Daniel Haziza and
                  Francisco Massa and
                  Alaaeldin El{-}Nouby and
                  Mido Assran and
                  Nicolas Ballas and
                  Wojciech Galuba and
                  Russell Howes and
                  Po{-}Yao Huang and
                  Shang{-}Wen Li and
                  Ishan Misra and
                  Michael Rabbat and
                  Vasu Sharma and
                  Gabriel Synnaeve and
                  Hu Xu and
                  Herv{\'{e}} J{\'{e}}gou and
                  Julien Mairal and
                  Patrick Labatut and
                  Armand Joulin and
                  Piotr Bojanowski},
  title        = {DINOv2: Learning Robust Visual Features without Supervision},
  journal      = {Trans. Mach. Learn. Res.},
  volume       = {2024},
  year         = {2024},
}

@article{what_pretrain,
  title={What Makes Pre-Trained Visual Representations Successful for Robust Manipulation?},
  author={Burns, Kaylee and Witzel, Zach and Hamid, Jubayer Ibn and Yu, Tianhe and Finn, Chelsea and Hausman, Karol},
  journal={arXiv preprint arXiv:2312.12444},
  year={2023}
}

@inproceedings{resnet,
  author       = {Kaiming He and
                  Xiangyu Zhang and
                  Shaoqing Ren and
                  Jian Sun},
  title        = {Deep Residual Learning for Image Recognition},
  booktitle    = {2016 {IEEE} Conference on Computer Vision and Pattern Recognition,
                  {CVPR} 2016, Las Vegas, NV, USA, June 27-30, 2016},
  pages        = {770--778},
  publisher    = {{IEEE} Computer Society},
  year         = {2016},
}

@inproceedings{perceiver,
  author       = {Andrew Jaegle and
                  Sebastian Borgeaud and
                  Jean{-}Baptiste Alayrac and
                  Carl Doersch and
                  Catalin Ionescu and
                  David Ding and
                  Skanda Koppula and
                  Daniel Zoran and
                  Andrew Brock and
                  Evan Shelhamer and
                  Olivier J. H{\'{e}}naff and
                  Matthew M. Botvinick and
                  Andrew Zisserman and
                  Oriol Vinyals and
                  Jo{\~{a}}o Carreira},
  title        = {Perceiver {IO:} {A} General Architecture for Structured Inputs {\&}
                  Outputs},
  booktitle    = {The Tenth International Conference on Learning Representations, {ICLR}
                  2022, Virtual Event, April 25-29, 2022},
  publisher    = {OpenReview.net},
  year         = {2022},
}

@inproceedings{umi,
	title={Universal Manipulation Interface: In-The-Wild Robot Teaching Without In-The-Wild Robots},
	author={Chi, Cheng and Xu, Zhenjia and Pan, Chuer and Cousineau, Eric and Burchfiel, Benjamin and Feng, Siyuan and Tedrake, Russ and Song, Shuran},
	booktitle={Proceedings of Robotics: Science and Systems (RSS)},
	year={2024}
}

@inproceedings{octo,
    title={Octo: An Open-Source Generalist Robot Policy},
    author = {{Octo Model Team} and Dibya Ghosh and Homer Walke and Karl Pertsch and Kevin Black and Oier Mees and Sudeep Dasari and Joey Hejna and Charles Xu and Jianlan Luo and Tobias Kreiman and {You Liang} Tan and Lawrence Yunliang Chen and Pannag Sanketi and Quan Vuong and Ted Xiao and Dorsa Sadigh and Chelsea Finn and Sergey Levine},
    booktitle = {Proceedings of Robotics: Science and Systems (RSS)},
    year = {2024},
}

@inproceedings{soft,
  author       = {Jianing Qian and
                  Anastasios Panagopoulos and
                  Dinesh Jayaraman},
  title        = {Recasting Generic Pretrained Vision Transformers As Object-Centric
                  Scene Encoders For Manipulation Policies},
  booktitle    = {{IEEE} International Conference on Robotics and Automation, {ICRA}
                  2024, Yokohama, Japan, May 13-17, 2024},
  pages        = {17544--17552},
  publisher    = {{IEEE}},
  year         = {2024},
}

@inproceedings{lora,
  author       = {Edward J. Hu and
                  Yelong Shen and
                  Phillip Wallis and
                  Zeyuan Allen{-}Zhu and
                  Yuanzhi Li and
                  Shean Wang and
                  Lu Wang and
                  Weizhu Chen},
  title        = {LoRA: Low-Rank Adaptation of Large Language Models},
  booktitle    = {The Tenth International Conference on Learning Representations, {ICLR}
                  2022, Virtual Event, April 25-29, 2022},
  publisher    = {OpenReview.net},
  year         = {2022},
}

@inproceedings{rt1,
  author       = {Anthony Brohan and
                  Noah Brown and
                  Justice Carbajal and
                  Yevgen Chebotar and
                  Joseph Dabis and
                  Chelsea Finn and
                  Keerthana Gopalakrishnan and
                  Karol Hausman and
                  Alexander Herzog and
                  Jasmine Hsu and
                  Julian Ibarz and
                  Brian Ichter and
                  Alex Irpan and
                  Tomas Jackson and
                  Sally Jesmonth and
                  Nikhil J. Joshi and
                  Ryan Julian and
                  Dmitry Kalashnikov and
                  Yuheng Kuang and
                  Isabel Leal and
                  Kuang{-}Huei Lee and
                  Sergey Levine and
                  Yao Lu and
                  Utsav Malla and
                  Deeksha Manjunath and
                  Igor Mordatch and
                  Ofir Nachum and
                  Carolina Parada and
                  Jodilyn Peralta and
                  Emily Perez and
                  Karl Pertsch and
                  Jornell Quiambao and
                  Kanishka Rao and
                  Michael S. Ryoo and
                  Grecia Salazar and
                  Pannag R. Sanketi and
                  Kevin Sayed and
                  Jaspiar Singh and
                  Sumedh Sontakke and
                  Austin Stone and
                  Clayton Tan and
                  Huong T. Tran and
                  Vincent Vanhoucke and
                  Steve Vega and
                  Quan Vuong and
                  Fei Xia and
                  Ted Xiao and
                  Peng Xu and
                  Sichun Xu and
                  Tianhe Yu and
                  Brianna Zitkovich},
  editor       = {Kostas E. Bekris and
                  Kris Hauser and
                  Sylvia L. Herbert and
                  Jingjin Yu},
  title        = {{RT-1:} Robotics Transformer for Real-World Control at Scale},
  booktitle    = {Robotics: Science and Systems XIX, Daegu, Republic of Korea, July
                  10-14, 2023},
  year         = {2023},
}

@inproceedings{rt2,
  author       = {Brianna Zitkovich and
                  Tianhe Yu and
                  Sichun Xu and
                  Peng Xu and
                  Ted Xiao and
                  Fei Xia and
                  Jialin Wu and
                  Paul Wohlhart and
                  Stefan Welker and
                  Ayzaan Wahid and
                  Quan Vuong and
                  Vincent Vanhoucke and
                  Huong T. Tran and
                  Radu Soricut and
                  Anikait Singh and
                  Jaspiar Singh and
                  Pierre Sermanet and
                  Pannag R. Sanketi and
                  Grecia Salazar and
                  Michael S. Ryoo and
                  Krista Reymann and
                  Kanishka Rao and
                  Karl Pertsch and
                  Igor Mordatch and
                  Henryk Michalewski and
                  Yao Lu and
                  Sergey Levine and
                  Lisa Lee and
                  Tsang{-}Wei Edward Lee and
                  Isabel Leal and
                  Yuheng Kuang and
                  Dmitry Kalashnikov and
                  Ryan Julian and
                  Nikhil J. Joshi and
                  Alex Irpan and
                  Brian Ichter and
                  Jasmine Hsu and
                  Alexander Herzog and
                  Karol Hausman and
                  Keerthana Gopalakrishnan and
                  Chuyuan Fu and
                  Pete Florence and
                  Chelsea Finn and
                  Kumar Avinava Dubey and
                  Danny Driess and
                  Tianli Ding and
                  Krzysztof Marcin Choromanski and
                  Xi Chen and
                  Yevgen Chebotar and
                  Justice Carbajal and
                  Noah Brown and
                  Anthony Brohan and
                  Montserrat Gonzalez Arenas and
                  Kehang Han},
  editor       = {Jie Tan and
                  Marc Toussaint and
                  Kourosh Darvish},
  title        = {{RT-2:} Vision-Language-Action Models Transfer Web Knowledge to Robotic
                  Control},
  booktitle    = {Conference on Robot Learning, CoRL 2023, 6-9 November 2023, Atlanta,
                  GA, {USA}},
  series       = {Proceedings of Machine Learning Research},
  volume       = {229},
  pages        = {2165--2183},
  publisher    = {{PMLR}},
  year         = {2023},
}

@inproceedings{mt-act,
  author       = {Homanga Bharadhwaj and
                  Jay Vakil and
                  Mohit Sharma and
                  Abhinav Gupta and
                  Shubham Tulsiani and
                  Vikash Kumar},
  title        = {RoboAgent: Generalization and Efficiency in Robot Manipulation via
                  Semantic Augmentations and Action Chunking},
  booktitle    = {{IEEE} International Conference on Robotics and Automation, {ICRA}
                  2024, Yokohama, Japan, May 13-17, 2024},
  pages        = {4788--4795},
  publisher    = {{IEEE}},
  year         = {2024},
}

@inproceedings{film,
  author       = {Ethan Perez and
                  Florian Strub and
                  Harm de Vries and
                  Vincent Dumoulin and
                  Aaron C. Courville},
  editor       = {Sheila A. McIlraith and
                  Kilian Q. Weinberger},
  title        = {FiLM: Visual Reasoning with a General Conditioning Layer},
  booktitle    = {Proceedings of the Thirty-Second {AAAI} Conference on Artificial Intelligence,
                  (AAAI-18), the 30th innovative Applications of Artificial Intelligence
                  (IAAI-18), and the 8th {AAAI} Symposium on Educational Advances in
                  Artificial Intelligence (EAAI-18), New Orleans, Louisiana, USA, February
                  2-7, 2018},
  pages        = {3942--3951},
  publisher    = {{AAAI} Press},
  year         = {2018},
}

@inproceedings{transformer,
  author       = {Ashish Vaswani and
                  Noam Shazeer and
                  Niki Parmar and
                  Jakob Uszkoreit and
                  Llion Jones and
                  Aidan N. Gomez and
                  Lukasz Kaiser and
                  Illia Polosukhin},
  editor       = {Isabelle Guyon and
                  Ulrike von Luxburg and
                  Samy Bengio and
                  Hanna M. Wallach and
                  Rob Fergus and
                  S. V. N. Vishwanathan and
                  Roman Garnett},
  title        = {Attention is All you Need},
  booktitle    = {Advances in Neural Information Processing Systems 30: Annual Conference
                  on Neural Information Processing Systems 2017, December 4-9, 2017,
                  Long Beach, CA, {USA}},
  pages        = {5998--6008},
  year         = {2017},
}

@article{droid,
  title={Droid: A large-scale in-the-wild robot manipulation dataset},
  author={Khazatsky, Alexander and Pertsch, Karl and Nair, Suraj and Balakrishna, Ashwin and Dasari, Sudeep and Karamcheti, Siddharth and Nasiriany, Soroush and Srirama, Mohan Kumar and Chen, Lawrence Yunliang and Ellis, Kirsty and others},
  journal={arXiv preprint arXiv:2403.12945},
  year={2024}
}

@inproceedings{unet,
  author       = {Olaf Ronneberger and
                  Philipp Fischer and
                  Thomas Brox},
  editor       = {Nassir Navab and
                  Joachim Hornegger and
                  William M. Wells III and
                  Alejandro F. Frangi},
  title        = {U-Net: Convolutional Networks for Biomedical Image Segmentation},
  booktitle    = {Medical Image Computing and Computer-Assisted Intervention - {MICCAI}
                  2015 - 18th International Conference Munich, Germany, October 5 -
                  9, 2015, Proceedings, Part {III}},
  series       = {Lecture Notes in Computer Science},
  volume       = {9351},
  pages        = {234--241},
  publisher    = {Springer},
  year         = {2015},
}

@inproceedings{sd,
  author       = {Robin Rombach and
                  Andreas Blattmann and
                  Dominik Lorenz and
                  Patrick Esser and
                  Bj{\"{o}}rn Ommer},
  title        = {High-Resolution Image Synthesis with Latent Diffusion Models},
  booktitle    = {{IEEE/CVF} Conference on Computer Vision and Pattern Recognition,
                  {CVPR} 2022, New Orleans, LA, USA, June 18-24, 2022},
  pages        = {10674--10685},
  publisher    = {{IEEE}},
  year         = {2022},
}

@inproceedings{robomimic,
  author       = {Ajay Mandlekar and
                  Danfei Xu and
                  Josiah Wong and
                  Soroush Nasiriany and
                  Chen Wang and
                  Rohun Kulkarni and
                  Li Fei{-}Fei and
                  Silvio Savarese and
                  Yuke Zhu and
                  Roberto Mart{\'{\i}}n{-}Mart{\'{\i}}n},
  editor       = {Aleksandra Faust and
                  David Hsu and
                  Gerhard Neumann},
  title        = {What Matters in Learning from Offline Human Demonstrations for Robot
                  Manipulation},
  booktitle    = {Conference on Robot Learning, 8-11 November 2021, London, {UK}},
  series       = {Proceedings of Machine Learning Research},
  volume       = {164},
  pages        = {1678--1690},
  publisher    = {{PMLR}},
  year         = {2021},
}

@article{rise,
  title={RISE: 3D Perception Makes Real-World Robot Imitation Simple and Effective},
  author={Wang, Chenxi and Fang, Hongjie and Fang, Hao-Shu and Lu, Cewu},
  journal={arXiv preprint arXiv:2404.12281},
  year={2024}
}

@inproceedings{rh20t,
  author       = {Haoshu Fang and
                  Hongjie Fang and
                  Zhenyu Tang and
                  Jirong Liu and
                  Chenxi Wang and
                  Junbo Wang and
                  Haoyi Zhu and
                  Cewu Lu},
  title        = {{RH20T:} {A} Comprehensive Robotic Dataset for Learning Diverse Skills
                  in One-Shot},
  booktitle    = {{IEEE} International Conference on Robotics and Automation, {ICRA}
                  2024, Yokohama, Japan, May 13-17, 2024},
  pages        = {653--660},
  publisher    = {{IEEE}},
  year         = {2024},
}

@inproceedings{vit,
  author       = {Alexey Dosovitskiy and
                  Lucas Beyer and
                  Alexander Kolesnikov and
                  Dirk Weissenborn and
                  Xiaohua Zhai and
                  Thomas Unterthiner and
                  Mostafa Dehghani and
                  Matthias Minderer and
                  Georg Heigold and
                  Sylvain Gelly and
                  Jakob Uszkoreit and
                  Neil Houlsby},
  title        = {An Image is Worth 16x16 Words: Transformers for Image Recognition
                  at Scale},
  booktitle    = {9th International Conference on Learning Representations, {ICLR} 2021,
                  Virtual Event, Austria, May 3-7, 2021},
  publisher    = {OpenReview.net},
  year         = {2021},
}

@article{maniwav,
  title={ManiWAV: Learning Robot Manipulation from In-the-Wild Audio-Visual Data},
  author={Liu, Zeyi and Chi, Cheng and Cousineau, Eric and Kuppuswamy, Naveen and Burchfiel, Benjamin and Song, Shuran},
  journal={arXiv preprint arXiv:2406.19464},
  year={2024}
}

@article{kat,
  title={Keypoint Action Tokens Enable In-Context Imitation Learning in Robotics},
  author={Di Palo, Norman and Johns, Edward},
  journal={arXiv preprint arXiv:2403.19578},
  year={2024}
}

@article{dexcap,
  title = {DexCap: Scalable and Portable Mocap Data Collection System for Dexterous Manipulation},
  author = {Wang, Chen and Shi, Haochen and Wang, Weizhuo and Zhang, Ruohan and Fei-Fei, Li and Liu, C. Karen},
  journal = {arXiv preprint arXiv:2403.07788},
  year = {2024}
}

@inproceedings{peract,
  author       = {Mohit Shridhar and
                  Lucas Manuelli and
                  Dieter Fox},
  editor       = {Karen Liu and
                  Dana Kulic and
                  Jeffrey Ichnowski},
  title        = {Perceiver-Actor: {A} Multi-Task Transformer for Robotic Manipulation},
  booktitle    = {Conference on Robot Learning, CoRL 2022, 14-18 December 2022, Auckland,
                  New Zealand},
  series       = {Proceedings of Machine Learning Research},
  volume       = {205},
  pages        = {785--799},
  publisher    = {{PMLR}},
  year         = {2022},
}

@inproceedings{emerging,
  author       = {Mathilde Caron and
                  Hugo Touvron and
                  Ishan Misra and
                  Herv{\'{e}} J{\'{e}}gou and
                  Julien Mairal and
                  Piotr Bojanowski and
                  Armand Joulin},
  title        = {Emerging Properties in Self-Supervised Vision Transformers},
  booktitle    = {2021 {IEEE/CVF} International Conference on Computer Vision, {ICCV}
                  2021, Montreal, QC, Canada, October 10-17, 2021},
  pages        = {9630--9640},
  publisher    = {{IEEE}},
  year         = {2021},
}

@inproceedings{oxe,
  title={Open X-Embodiment: Robotic Learning Datasets and RT-X Models},
  author={Open X-Embodiment Collaboration and O’Neill, Abby and Rehman, Abdul and Maddukuri, Abhiram and Gupta, Abhishek and Padalkar, Abhishek and Lee, Abraham and Pooley, Acorn and Gupta, Agrim and Mandlekar, Ajay and Jain, Ajinkya and others},
  booktitle={2024 IEEE International Conference on Robotics and Automation (ICRA)},
  pages={6892--6903},
  year={2024},
  organization={IEEE}
}

@misc{flexiv,
    title={Flexiv Rizon Robot},
    url={https://www.flexiv.com/product/rizon},
    year={2024},
    month={9}
}

@misc{dahuan,
    title={Dahuan AG Series Gripper},
    url={https://en.dh-robotics.com/product/ag},
    year={2024},
    month={9}
}

@misc{realsense,
    title={Intel RealSense Depth Camera D435},
    url={https://www.intelrealsense.com/depth-camera-d435},
    year={2024},
    month={9}
}

@misc{realsense-415,
    title={Intel RealSense Depth Camera D415},
    url={https://www.intelrealsense.com/depth-camera-d415},
    year={2024},
    month={9}
}

@misc{sigma7,
    title={Force Dimension - sigma.7},
    url={https://www.forcedimension.com/products/sigma},
    year={2024},
    month={9}
}

@inproceedings{gnfactor,
  author       = {Yanjie Ze and
                  Ge Yan and
                  Yueh{-}Hua Wu and
                  Annabella Macaluso and
                  Yuying Ge and
                  Jianglong Ye and
                  Nicklas Hansen and
                  Li Erran Li and
                  Xiaolong Wang},
  editor       = {Jie Tan and
                  Marc Toussaint and
                  Kourosh Darvish},
  title        = {GNFactor: Multi-Task Real Robot Learning with Generalizable Neural
                  Feature Fields},
  booktitle    = {Conference on Robot Learning, CoRL 2023, 6-9 November 2023, Atlanta,
                  GA, {USA}},
  series       = {Proceedings of Machine Learning Research},
  volume       = {229},
  pages        = {284--301},
  publisher    = {{PMLR}},
  year         = {2023},
}

@inproceedings{dp3d,
	title={3D Diffusion Policy: Generalizable Visuomotor Policy Learning via Simple 3D Representations},
	author={Yanjie Ze and Gu Zhang and Kangning Zhang and Chenyuan Hu and Muhan Wang and Huazhe Xu},
	booktitle={Proceedings of Robotics: Science and Systems (RSS)},
	year={2024}
}

@inproceedings{act3d,
  author       = {Th{\'{e}}ophile Gervet and
                  Zhou Xian and
                  Nikolaos Gkanatsios and
                  Katerina Fragkiadaki},
  editor       = {Jie Tan and
                  Marc Toussaint and
                  Kourosh Darvish},
  title        = {Act3D: 3D Feature Field Transformers for Multi-Task Robotic Manipulation},
  booktitle    = {Conference on Robot Learning, CoRL 2023, 6-9 November 2023, Atlanta,
                  GA, {USA}},
  series       = {Proceedings of Machine Learning Research},
  volume       = {229},
  pages        = {3949--3965},
  publisher    = {{PMLR}},
  year         = {2023},
}

@inproceedings{sgr,
  author       = {Tong Zhang and
                  Yingdong Hu and
                  Hanchen Cui and
                  Hang Zhao and
                  Yang Gao},
  editor       = {Jie Tan and
                  Marc Toussaint and
                  Kourosh Darvish},
  title        = {A Universal Semantic-Geometric Representation for Robotic Manipulation},
  booktitle    = {Conference on Robot Learning, CoRL 2023, 6-9 November 2023, Atlanta,
                  GA, {USA}},
  series       = {Proceedings of Machine Learning Research},
  volume       = {229},
  pages        = {3342--3363},
  publisher    = {{PMLR}},
  year         = {2023},
}

@inproceedings{vc1,
  author       = {Arjun Majumdar and
                  Karmesh Yadav and
                  Sergio Arnaud and
                  Yecheng Jason Ma and
                  Claire Chen and
                  Sneha Silwal and
                  Aryan Jain and
                  Vincent{-}Pierre Berges and
                  Tingfan Wu and
                  Jay Vakil and
                  Pieter Abbeel and
                  Jitendra Malik and
                  Dhruv Batra and
                  Yixin Lin and
                  Oleksandr Maksymets and
                  Aravind Rajeswaran and
                  Franziska Meier},
  editor       = {Alice Oh and
                  Tristan Naumann and
                  Amir Globerson and
                  Kate Saenko and
                  Moritz Hardt and
                  Sergey Levine},
  title        = {Where are we in the search for an Artificial Visual Cortex for Embodied
                  Intelligence?},
  booktitle    = {Advances in Neural Information Processing Systems 36: Annual Conference
                  on Neural Information Processing Systems 2023, NeurIPS 2023, New Orleans,
                  LA, USA, December 10 - 16, 2023},
  year         = {2023},
}

@article{theia,
  title={Theia: Distilling Diverse Vision Foundation Models for Robot Learning},
  author={Shang, Jinghuan and Schmeckpeper, Karl and May, Brandon B and Minniti, Maria Vittoria and Kelestemur, Tarik and Watkins, David and Herlant, Laura},
  journal={arXiv preprint arXiv:2407.20179},
  year={2024}
}

@inproceedings{moco,
  author       = {Xinlei Chen and
                  Saining Xie and
                  Kaiming He},
  title        = {An Empirical Study of Training Self-Supervised Vision Transformers},
  booktitle    = {2021 {IEEE/CVF} International Conference on Computer Vision, {ICCV}
                  2021, Montreal, QC, Canada, October 10-17, 2021},
  pages        = {9620--9629},
  publisher    = {{IEEE}},
  year         = {2021},
}

@inproceedings{clip,
  author       = {Alec Radford and
                  Jong Wook Kim and
                  Chris Hallacy and
                  Aditya Ramesh and
                  Gabriel Goh and
                  Sandhini Agarwal and
                  Girish Sastry and
                  Amanda Askell and
                  Pamela Mishkin and
                  Jack Clark and
                  Gretchen Krueger and
                  Ilya Sutskever},
  editor       = {Marina Meila and
                  Tong Zhang},
  title        = {Learning Transferable Visual Models From Natural Language Supervision},
  booktitle    = {Proceedings of the 38th International Conference on Machine Learning,
                  {ICML} 2021, 18-24 July 2021, Virtual Event},
  series       = {Proceedings of Machine Learning Research},
  volume       = {139},
  pages        = {8748--8763},
  publisher    = {{PMLR}},
  year         = {2021},
}

@inproceedings{bc,
  author       = {Dean Pomerleau},
  editor       = {David S. Touretzky},
  title        = {{ALVINN:} An Autonomous Land Vehicle in a Neural Network},
  booktitle    = {Advances in Neural Information Processing Systems 1, {[NIPS} Conference,
                  Denver, Colorado, USA, 1988]},
  pages        = {305--313},
  publisher    = {Morgan Kaufmann},
  year         = {1988},
}

@inproceedings{vinn,
  author       = {Jyothish Pari and
                  Nur Muhammad (Mahi) Shafiullah and
                  Sridhar Pandian Arunachalam and
                  Lerrel Pinto},
  editor       = {Kris Hauser and
                  Dylan A. Shell and
                  Shoudong Huang},
  title        = {The Surprising Effectiveness of Representation Learning for Visual
                  Imitation},
  booktitle    = {Robotics: Science and Systems XVIII, New York City, NY, USA, June
                  27 - July 1, 2022},
  year         = {2022},
}

@inproceedings{mvp,
  author       = {Ilija Radosavovic and
                  Tete Xiao and
                  Stephen James and
                  Pieter Abbeel and
                  Jitendra Malik and
                  Trevor Darrell},
  editor       = {Karen Liu and
                  Dana Kulic and
                  Jeffrey Ichnowski},
  title        = {Real-World Robot Learning with Masked Visual Pre-training},
  booktitle    = {Conference on Robot Learning, CoRL 2022, 14-18 December 2022, Auckland,
                  New Zealand},
  series       = {Proceedings of Machine Learning Research},
  volume       = {205},
  pages        = {416--426},
  publisher    = {{PMLR}},
  year         = {2022},
}

@inproceedings{same,
  author       = {Junjie Zhang and
                  Chenjia Bai and
                  Haoran He and
                  Zhigang Wang and
                  Bin Zhao and
                  Xiu Li and
                  Xuelong Li},
  title        = {{SAM-E:} Leveraging Visual Foundation Model with Sequence Imitation
                  for Embodied Manipulation},
  booktitle    = {Forty-first International Conference on Machine Learning, {ICML} 2024,
                  Vienna, Austria, July 21-27, 2024},
  publisher    = {OpenReview.net},
  year         = {2024},
}

@inproceedings{spawnnet,
  author       = {Xingyu Lin and
                  John So and
                  Sashwat Mahalingam and
                  Fangchen Liu and
                  Pieter Abbeel},
  title        = {SpawnNet: Learning Generalizable Visuomotor Skills from Pre-trained
                  Network},
  booktitle    = {{IEEE} International Conference on Robotics and Automation, {ICRA}
                  2024, Yokohama, Japan, May 13-17, 2024},
  pages        = {4781--4787},
  publisher    = {{IEEE}},
  year         = {2024},
}

@inproceedings{sam,
  author       = {Alexander Kirillov and
                  Eric Mintun and
                  Nikhila Ravi and
                  Hanzi Mao and
                  Chlo{\'{e}} Rolland and
                  Laura Gustafson and
                  Tete Xiao and
                  Spencer Whitehead and
                  Alexander C. Berg and
                  Wan{-}Yen Lo and
                  Piotr Doll{\'{a}}r and
                  Ross B. Girshick},
  title        = {Segment Anything},
  booktitle    = {{IEEE/CVF} International Conference on Computer Vision, {ICCV} 2023,
                  Paris, France, October 1-6, 2023},
  pages        = {3992--4003},
  publisher    = {{IEEE}},
  year         = {2023},
}

@inproceedings{byol,
  author       = {Jean{-}Bastien Grill and
                  Florian Strub and
                  Florent Altch{\'{e}} and
                  Corentin Tallec and
                  Pierre H. Richemond and
                  Elena Buchatskaya and
                  Carl Doersch and
                  Bernardo {\'{A}}vila Pires and
                  Zhaohan Guo and
                  Mohammad Gheshlaghi Azar and
                  Bilal Piot and
                  Koray Kavukcuoglu and
                  R{\'{e}}mi Munos and
                  Michal Valko},
  editor       = {Hugo Larochelle and
                  Marc'Aurelio Ranzato and
                  Raia Hadsell and
                  Maria{-}Florina Balcan and
                  Hsuan{-}Tien Lin},
  title        = {Bootstrap Your Own Latent - {A} New Approach to Self-Supervised Learning},
  booktitle    = {Advances in Neural Information Processing Systems 33: Annual Conference
                  on Neural Information Processing Systems 2020, NeurIPS 2020, December
                  6-12, 2020, virtual},
  year         = {2020},
}

@inproceedings{mae,
  author       = {Kaiming He and
                  Xinlei Chen and
                  Saining Xie and
                  Yanghao Li and
                  Piotr Doll{\'{a}}r and
                  Ross B. Girshick},
  title        = {Masked Autoencoders Are Scalable Vision Learners},
  booktitle    = {{IEEE/CVF} Conference on Computer Vision and Pattern Recognition,
                  {CVPR} 2022, New Orleans, LA, USA, June 18-24, 2022},
  pages        = {15979--15988},
  publisher    = {{IEEE}},
  year         = {2022},
}

@inproceedings{ego4d,
  author       = {Kristen Grauman and
                  Andrew Westbury and
                  Eugene Byrne and
                  Zachary Chavis and
                  Antonino Furnari and
                  Rohit Girdhar and
                  Jackson Hamburger and
                  Hao Jiang and
                  Miao Liu and
                  Xingyu Liu and
                  Miguel Martin and
                  Tushar Nagarajan and
                  Ilija Radosavovic and
                  Santhosh Kumar Ramakrishnan and
                  Fiona Ryan and
                  Jayant Sharma and
                  Michael Wray and
                  Mengmeng Xu and
                  Eric Zhongcong Xu and
                  Chen Zhao and
                  Siddhant Bansal and
                  Dhruv Batra and
                  Vincent Cartillier and
                  Sean Crane and
                  Tien Do and
                  Morrie Doulaty and
                  Akshay Erapalli and
                  Christoph Feichtenhofer and
                  Adriano Fragomeni and
                  Qichen Fu and
                  Abrham Gebreselasie and
                  Cristina Gonz{\'{a}}lez and
                  James Hillis and
                  Xuhua Huang and
                  Yifei Huang and
                  Wenqi Jia and
                  Weslie Khoo and
                  J{\'{a}}chym Kol{\'{a}}r and
                  Satwik Kottur and
                  Anurag Kumar and
                  Federico Landini and
                  Chao Li and
                  Yanghao Li and
                  Zhenqiang Li and
                  Karttikeya Mangalam and
                  Raghava Modhugu and
                  Jonathan Munro and
                  Tullie Murrell and
                  Takumi Nishiyasu and
                  Will Price and
                  Paola Ruiz Puentes and
                  Merey Ramazanova and
                  Leda Sari and
                  Kiran Somasundaram and
                  Audrey Southerland and
                  Yusuke Sugano and
                  Ruijie Tao and
                  Minh Vo and
                  Yuchen Wang and
                  Xindi Wu and
                  Takuma Yagi and
                  Ziwei Zhao and
                  Yunyi Zhu and
                  Pablo Arbel{\'{a}}ez and
                  David Crandall and
                  Dima Damen and
                  Giovanni Maria Farinella and
                  Christian Fuegen and
                  Bernard Ghanem and
                  Vamsi Krishna Ithapu and
                  C. V. Jawahar and
                  Hanbyul Joo and
                  Kris Kitani and
                  Haizhou Li and
                  Richard A. Newcombe and
                  Aude Oliva and
                  Hyun Soo Park and
                  James M. Rehg and
                  Yoichi Sato and
                  Jianbo Shi and
                  Mike Zheng Shou and
                  Antonio Torralba and
                  Lorenzo Torresani and
                  Mingfei Yan and
                  Jitendra Malik},
  title        = {Ego4D: Around the World in 3, 000 Hours of Egocentric Video},
  booktitle    = {{IEEE/CVF} Conference on Computer Vision and Pattern Recognition,
                  {CVPR} 2022, New Orleans, LA, USA, June 18-24, 2022},
  pages        = {18973--18990},
  publisher    = {{IEEE}},
  year         = {2022},
}

@article{epickitchen,
  author       = {Dima Damen and
                  Hazel Doughty and
                  Giovanni Maria Farinella and
                  Antonino Furnari and
                  Evangelos Kazakos and
                  Jian Ma and
                  Davide Moltisanti and
                  Jonathan Munro and
                  Toby Perrett and
                  Will Price and
                  Michael Wray},
  title        = {Rescaling Egocentric Vision: Collection, Pipeline and Challenges for
                  {EPIC-KITCHENS-100}},
  journal      = {Int. J. Comput. Vis.},
  volume       = {130},
  number       = {1},
  pages        = {33--55},
  year         = {2022},
}

@inproceedings{sth-sth,
  author       = {Raghav Goyal and
                  Samira Ebrahimi Kahou and
                  Vincent Michalski and
                  Joanna Materzynska and
                  Susanne Westphal and
                  Heuna Kim and
                  Valentin Haenel and
                  Ingo Fr{\"{u}}nd and
                  Peter Yianilos and
                  Moritz Mueller{-}Freitag and
                  Florian Hoppe and
                  Christian Thurau and
                  Ingo Bax and
                  Roland Memisevic},
  title        = {The "Something Something" Video Database for Learning and Evaluating
                  Visual Common Sense},
  booktitle    = {{IEEE} International Conference on Computer Vision, {ICCV} 2017, Venice,
                  Italy, October 22-29, 2017},
  pages        = {5843--5851},
  publisher    = {{IEEE} Computer Society},
  year         = {2017},
}

@inproceedings{doh100,
  author       = {Dandan Shan and
                  Jiaqi Geng and
                  Michelle Shu and
                  David F. Fouhey},
  title        = {Understanding Human Hands in Contact at Internet Scale},
  booktitle    = {2020 {IEEE/CVF} Conference on Computer Vision and Pattern Recognition,
                  {CVPR} 2020, Seattle, WA, USA, June 13-19, 2020},
  pages        = {9866--9875},
  publisher    = {Computer Vision Foundation / {IEEE}},
  year         = {2020},
}

@article{anygrasp,
  author       = {Haoshu Fang and
                  Chenxi Wang and
                  Hongjie Fang and
                  Minghao Gou and
                  Jirong Liu and
                  Hengxu Yan and
                  Wenhai Liu and
                  Yichen Xie and
                  Cewu Lu},
  title        = {AnyGrasp: Robust and Efficient Grasp Perception in Spatial and Temporal
                  Domains},
  journal      = {{IEEE} Trans. Robotics},
  volume       = {39},
  number       = {5},
  pages        = {3929--3945},
  year         = {2023},
}

@inproceedings{dexgrasp,
  author       = {Weikang Wan and
                  Haoran Geng and
                  Yun Liu and
                  Zikang Shan and
                  Yaodong Yang and
                  Li Yi and
                  He Wang},
  title        = {UniDexGrasp++: Improving Dexterous Grasping Policy Learning via Geometry-aware
                  Curriculum and Iterative Generalist-Specialist Learning},
  booktitle    = {{IEEE/CVF} International Conference on Computer Vision, {ICCV} 2023,
                  Paris, France, October 1-6, 2023},
  pages        = {3868--3879},
  publisher    = {{IEEE}},
  year         = {2023},
}

@inproceedings{bridge-v2,
  author       = {Homer Rich Walke and
                  Kevin Black and
                  Tony Z. Zhao and
                  Quan Vuong and
                  Chongyi Zheng and
                  Philippe Hansen{-}Estruch and
                  Andre Wang He and
                  Vivek Myers and
                  Moo Jin Kim and
                  Max Du and
                  Abraham Lee and
                  Kuan Fang and
                  Chelsea Finn and
                  Sergey Levine},
  editor       = {Jie Tan and
                  Marc Toussaint and
                  Kourosh Darvish},
  title        = {BridgeData {V2:} {A} Dataset for Robot Learning at Scale},
  booktitle    = {Conference on Robot Learning, CoRL 2023, 6-9 November 2023, Atlanta,
                  GA, {USA}},
  series       = {Proceedings of Machine Learning Research},
  volume       = {229},
  pages        = {1723--1736},
  publisher    = {{PMLR}},
  year         = {2023}
}




\clearpage
\appendix

\subsection{Implementation Details}
\label{appendix:impl}

We provide detailed information about the architecture of \model~and our implementations of the selected baselines DP~\cite{dp} and RISE~\cite{rise}.
Main hyper-parameters used for training and evaluation are presented in Tab.~\ref{tab:comp-method} for comparison.

\subsubsection{CAGE} Our model uses \texttt{DINOv2-large}~\cite{dino} as the vision encoder backbone. In order to preserve its rich semantic knowledge learned from pre-training, the weights of DINOv2 are kept frozen and we employ LoRA~\cite{lora} on the projector matrices of the attention modules for parameter-efficient fine-tuning. LoRA rank is set to 16 and a dropout rate of 0.1 is applied to avoid over-fitting.
We set the observation horizon $T_o=4$ based on the ablation results (Tab.~\ref{tab:Ho} in \S\ref{sec:ablation}).
The images are first resized to $256\times256$ and then cropped to $224\times224$ as observation inputs.
The vision encoder outputs $L=256$ tokens of dimension $D = 1024$ for each frame. These tokens are first projected into $D_\text{token} = 512$. Then, $N_c\times T_o\times L=2\times 4 \times 256 = 2048$ tokens are compressed into $T_o=4$ tokens via our proposed \textit{Causal Observation Perceiver}.
The perceiver is composed of a cross-attention module followed by self-attention modules with causal masks:  $T_o$ learnable tokens are cross-attended to concatenated observation tokens using a causal mask for each camera;  the output tokens are then passed into a sequence of $N=4$ causal self-attention modules for further processing. Dropout layers with $p=0.1$ are also applied on each attention block for a more robust feature extraction.
Modified from the pure CNN-based noise prediction network~\cite{dp}, we append a \textit{Cross-Attention Block} to each \textit{Residual Block} for separate conditioning of observation and timestep. We adopt the default setting of a 3-layer UNet with increased channel dimensions 256-512-1024 as suggested in~\cite{dp}. For the data augmentation, only the color jitter is included in $L_0$ and $L_1$ generalization evaluations, and additional geometric augmentation (\S\ref{sec:ablation}) is applied in $L_2$ evaluations for better overall performance.

\subsubsection{DP} The CNN-based Diffusion Policy is implemented following \href{https://github.com/real-stanford/diffusion_policy/}{the official implementation}, except that the vision encoder is upgraded from \texttt{ResNet-18} to \texttt{ResNet-50} for a higher capacity in visual expressiveness. 
Given that DP performs better with a shorter observation horizon, we keep $T_o = 2$ and $T_a = 16$. All other hyper-parameters are aligned with those of \model~to facilitate direct comparisons. The augmentations also follows the \model~implementations. Moreover, a strictly aligned $T_o = 4$ and $T_a = 20$ version of DP is included in Tab.~\ref{tab:ablation} (4th row) for the ablation study.

\subsubsection{RISE} The configuration of RISE remains the same as \href{https://github.com/rise-policy/rise}{the official implementation}, except that we enable the color jitter augmentation option during training for fair generalization comparisons. Following their original settings, RISE only takes the point cloud from a single fixed camera with observation horizon $T_o=1$ as inputs, and outputs continuous actions for $T_a = 20$ steps. RISE adopts absolute action representations in the camera coordinate.

\begin{table*}
\centering
\renewcommand{\arraystretch}{1.2}
\begin{tabular}{c|ccc}
\toprule
    & CAGE \textit{(ours)} & DP~\cite{dp} & RISE~\cite{rise} \\ 
\midrule
Observation Horizon $T_o$ & 4          & 2            & 1                \\
Action Horizon $T_a$      & 20         & 16           & 20               \\
Proprioception         & \checkmark    & $\times$     & $\times$ \\
Camera Setting & 1 fixed + 1 in-hand & 1 fixed + 1 in-hand & 1 fixed \\
Observation Type       & RGB           & RGB               & RGB-D       \\
Observation Resolution & $224\times224$ (image) & $224\times224$ (image) & $1280\times720$ (point cloud) \\
\cmidrule(lr){1-4}
Vision Encoder & 
    DINOv2-large\begin{tabular}{c}\hspace{-0.29cm}\vspace{-0.06cm}\emojifreeze\end{tabular}\hspace{-0.15cm}\textit{w.} LoRA\begin{tabular}{c}\hspace{-0.3cm}\vspace{-0.05cm}\emojifire\end{tabular}\hspace{-0.2cm} &
    ResNet-50\begin{tabular}{c}\hspace{-0.3cm}\vspace{-0.05cm}\emojifire\end{tabular}\hspace{-0.2cm} &
    Sparse 3D Encoder\begin{tabular}{c}\hspace{-0.3cm}\vspace{-0.05cm}\emojifire\end{tabular}\hspace{-0.2cm} \\
Observation Perceiver & \checkmark & \diagbox[dir=SW]{}{} & \diagbox[dir=SW]{}{} \\
UNet of Diffusion Head & 3 layers (256-512-1024) & 3 layers (256-512-1024) & 2 layers (256-512) \\
\cmidrule(lr){1-4}
Denoising Steps & 16        & 16        & 16          \\
Inference Time  & 280ms     & 100ms     & 130ms       \\
Temporal Ensemble~\cite{act} & \checkmark\ (first 12 steps) & \checkmark\ (first 12 steps) & \checkmark\ (all $T_a$ steps)\\
Control Type    & parallel  & parallel  & blocking    \\
Control Frequency   & 10Hz  & 10Hz  & \diagbox[dir=SW]{}{} \\
Prediction Frequency & every 0.35s & every 0.35s & every 4 steps \\
\bottomrule
\end{tabular}%
\caption{\textbf{Detailed Comparison of All Policies.} \begin{tabular}{c}\hspace{-0.29cm}\vspace{-0.06cm}\emojifreeze\end{tabular}\hspace{-0.15cm}/\begin{tabular}{c}\hspace{-0.3cm}\vspace{-0.05cm}\emojifire\end{tabular}\hspace{-0.2cm} denotes frozen\,/\,trainable weights respectively.}
\label{tab:comp-method}
\vspace{-0.4cm}
\end{table*}

\subsection{Task Details}

\begin{figure}[t]
    \centering
    \includegraphics[width=0.85\columnwidth]{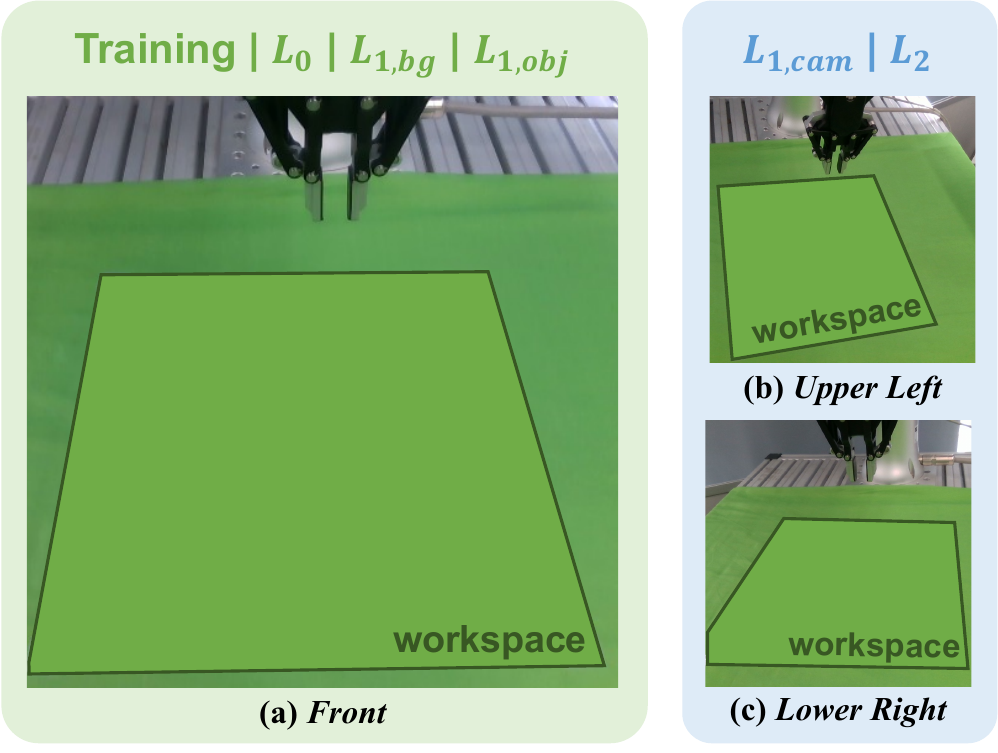}
    \caption{\textbf{Observation Images from Different Cameras.} The front camera is used as the fixed camera for training and camera-in-domain experiments ($L_0$, $L_{1,bg}$ and $L_{1,obj}$). The upper-left and lower-right cameras are utilized for camera-out-of-domain generalization experiments ($L_{1,cam}$ and $L_2$). Both alternative viewpoints introduce a considerable amount of visual distortion.}
    \label{fig:camera-views}
    \vspace{-0.4cm}
\end{figure}

\begin{figure*}
    \centering
    \includegraphics[width=0.9\textwidth]{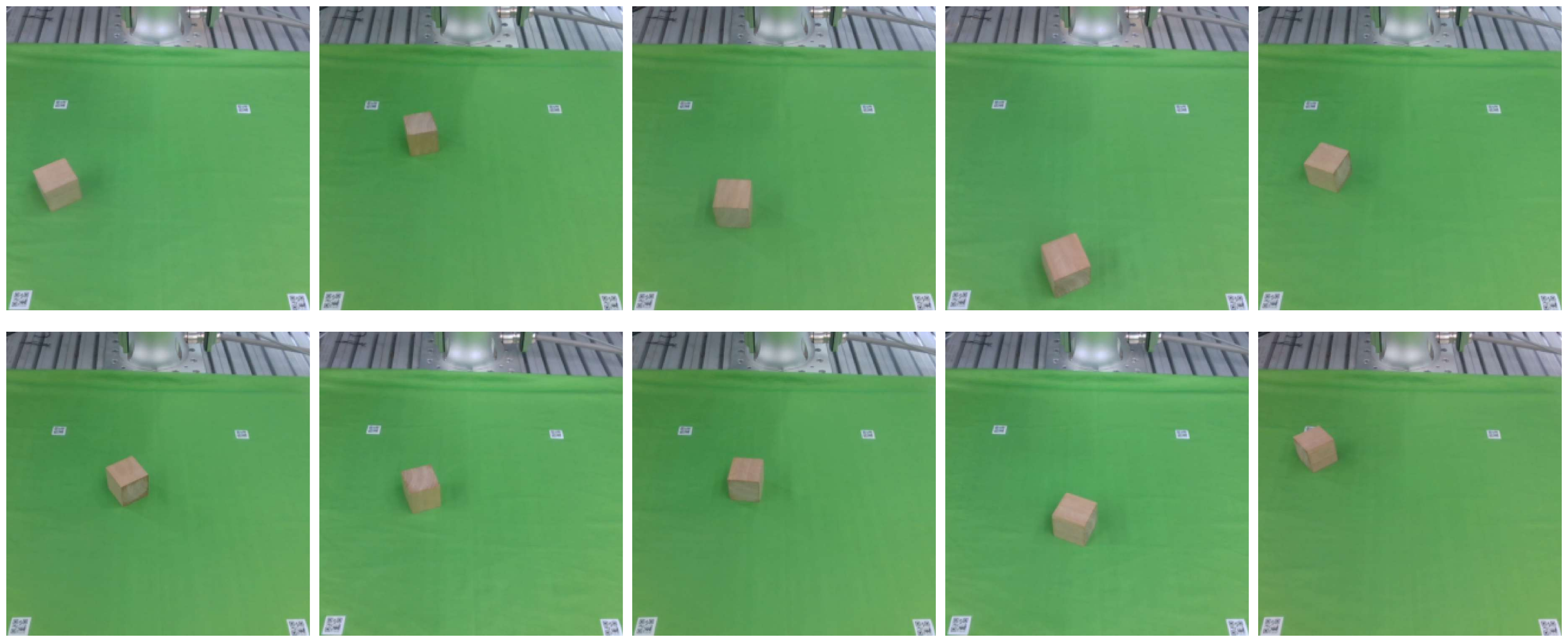}
    \caption{\textbf{Sample Initializations of the \textit{Transport} Task}. The block is randomly initialized on the left part of the workspace.}
    \label{fig:block-tasks}
\end{figure*}

\begin{figure*}
    \centering
    \includegraphics[width=0.9\textwidth]{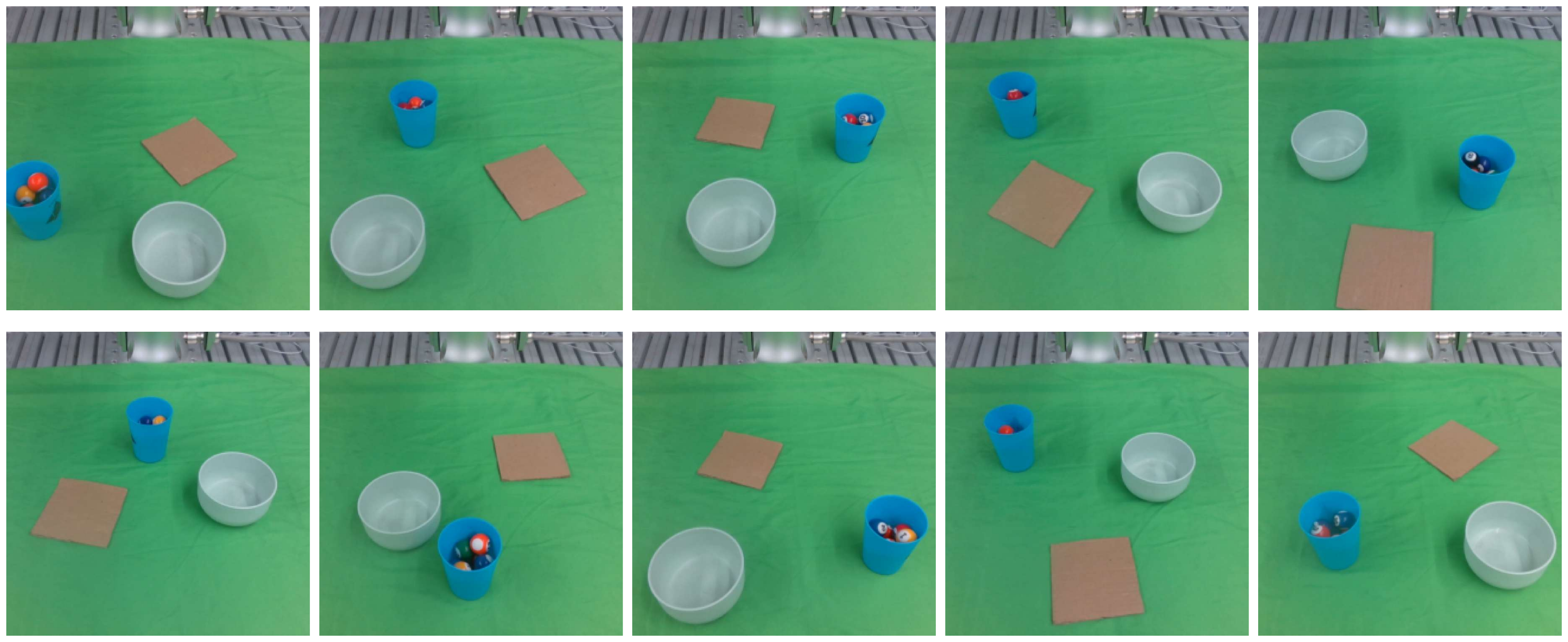}
    \caption{\textbf{Sample Initializations of the \textit{Dumping} Task}. The cup (with 10 balls in it), the bowl, and the cardboard as the placement target area are randomly initialized on the workspace.}
    \label{fig:balls-tasks}
\end{figure*}

\begin{figure*}
    \centering
    \includegraphics[width=0.9\textwidth]{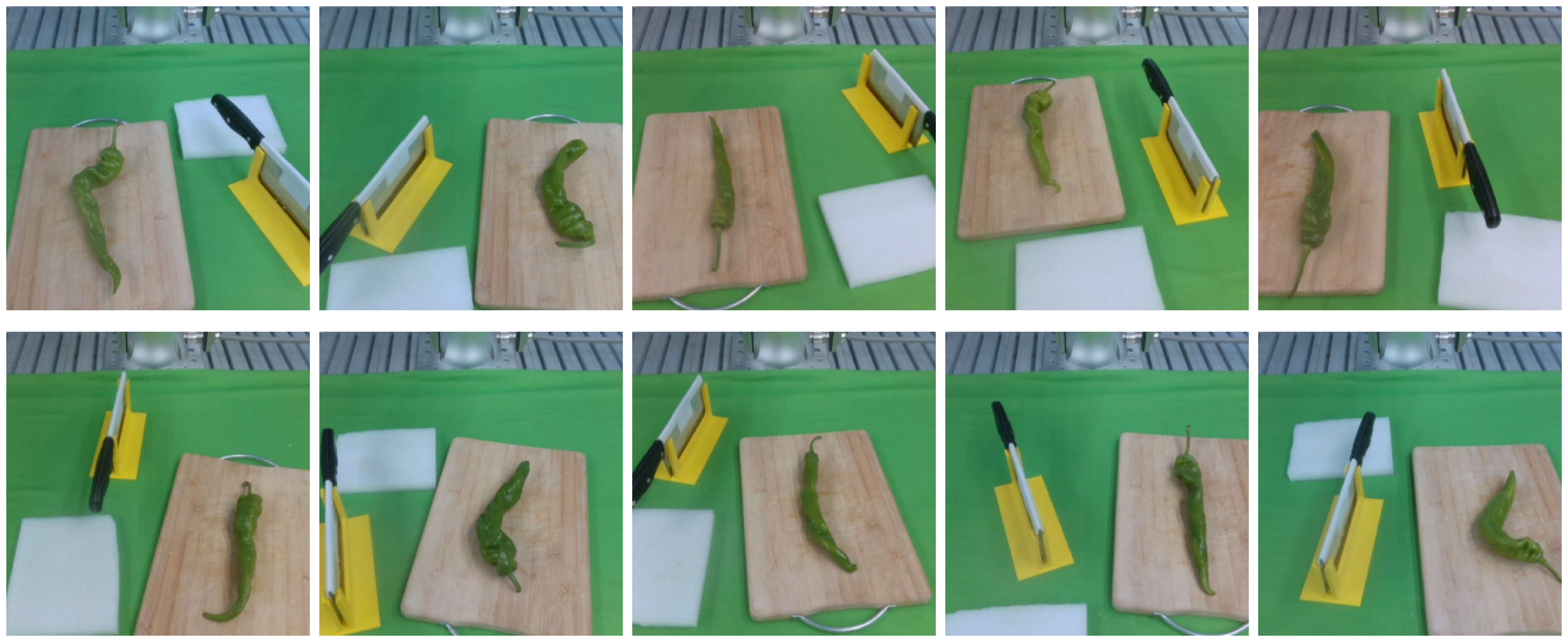}
    \caption{\textbf{Sample Initializations of the \textit{Chopping} Task.} A kitchen knife (with its base), a cutting board (with chili pepper on it), and a foam pad used for placement are randomly initialized on the workspace.}
    \label{fig:chop-tasks}
\end{figure*}

We design three types of tasks, each corresponding to a different aspect of the robotic manipulation problem.
\begin{itemize}
    \item \textbf{Pick-and-place task}: The pick-and-place task represents the most fundamental and widely utilized operation in robotic control. 
    The \textbf{\textit{Transport}} task (samples shown in Fig.~\ref{fig:block-tasks}) consists of the robotic arm first picking up the object on the left half of the workspace and dropping it to the right.
    This task necessitates the robot's capacity to accurately locate and \underline{reach} the target object, precisely \underline{grasp} it with an appropriate pose, and \underline{drop} it in the designated location.
    \textit{We use the \textbf{Transport} task to assess the policy's generalization ability for the elementary skills.}
    
    \item \textbf{6-DoF task}: By unlocking two more degrees of freedom in rotation, 6-DoF actions can greatly facilitate the robot's interaction with objects.
    The \textbf{\textit{Dumping}} task (samples shown in Fig.~\ref{fig:balls-tasks}) can be decomposed into 3 sub-tasks with rotations: \underline{grasp} the plastic cup horizontally; move it over the bowl and \underline{pour} all 10 balls into the bowl; \underline{place} the cup vertically inside the target area.
    During the prediction, the policy must determine if additional rotation is required based on the observation inputs. This presents a significant challenge for the model in terms of understanding its own state within the environment.
    \textit{We design the \textbf{Dumping} task to test whether the policy can understand its own state within the environment and produce coherent rotations against visual variations.}
    
    \item \textbf{Long-horizon task}: In many scenarios, a single task is composed of several repetitive small tasks that often have to be executed in order. This underscores the robustness and flexibility of the policy during the interaction with the environment.
    The objective of the \textbf{\textit{Chopping}} task (samples shown in Fig.~\ref{fig:chop-tasks}) is to \underline{grasp} properly the kitchen knife, \underline{chop} the pepper on the cutting board into several pieces, and \underline{place} the knife safely on the foam pad. The robotic arm is required to conduct multiple chopping evenly and progressively.
    \textit{We employ the \textbf{Chopping} task to evaluate the task planning and error recovery abilities of the model in out-of-distribution scenes.}
\end{itemize}

\subsection{Evaluation Details}

In order to evaluate the generalization performance of each policy, we devise distribution shifts on three principal visual factors that are particularly susceptible to variation upon deployment in the real world:
\begin{itemize}
\item \textbf{Background}: The original green background is changed to a white leather-textured one or a tablecloth with light and dark geometric patterns depending on the task.

\item \textbf{Object}: We modify the target object's color and geometry to evaluate the object-level generalizability of manipulation policies. For the \textbf{\textit{Dumping}} and the \textbf{\textit{Chopping}} tasks, in addition to the change of the target object (bowl in \textbf{\textit{Dumping}} and pepper in \textbf{\textit{Chopping}}), we also alter the color of the cardboard (in \textbf{\textit{Dumping}}) or the foam pad (in \textbf{\textit{Chopping}}). In the \textbf{\textit{Transport}} task, we additionally test the policies using a football, which poses a serious challenge due to the huge difference in geometry and the need for strong self-correction skills following unsuccessful attempts.

\item \textbf{Camera View}: We introduce a large visual distortions by displacing the camera approximately 16cm away from the original position. Additionally, the original fixed Intel RealSense D435 camera~\cite{realsense} is replaced with an Intel RealSense D415 camera~\cite{realsense-415} for a more challenging setup. The images from different cameras are shown in Fig.~\ref{fig:camera-views}, from which we can observe a significant view discrepancy between training and evaluation cameras.
\end{itemize}

Therefore, we define three levels of generalization tests:
\begin{itemize}
    \item \textbf{\textit{L}$_\mathbf{0}$ generalization} is conducted in the same environment seen during training. We evaluate the most basic ability of the policy to learn the task from demonstrations.
    \item \textbf{\textit{L}$_\mathbf{1}$ generalization} is conducted in a similar environment with the variation in one of the three visual aspects, which we refer to as $L_{1, bg}$, $L_{1, obj}$ and $L_{1, cam}$ respectively. We evaluate the robustness of the policy under moderate distribution shifts.
    \item \textbf{\textit{L}$_\mathbf{2}$ generalization} is conducted in a totally unseen environment with variations on all three aspects. We evaluate the capability of the policy to complete the learned task in out-of-domain scenes.
\end{itemize} 

\begin{figure}[htbp]
    \vspace{-0.4cm}
    \centering
    \includegraphics[width=0.8\columnwidth]{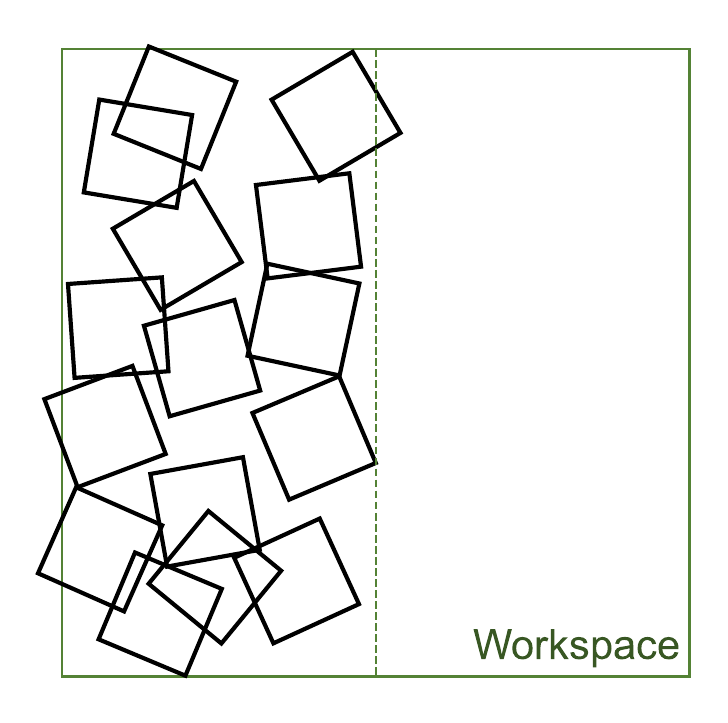}
    \vspace{-0.3cm}
    \caption{\textbf{Evaluation Initialization for the \textit{Transport} Task.} The target block is placed at each designated location for each evaluation round.}
    \label{fig:eval-pos}
    \vspace{-0.4cm}
\end{figure}

Moreover, we use a script\footnote{The script for generating random initial object positions is also provided in our code repository.} to generate randomly and uniformly distributed evaluation positions \textit{beforehand} for each task, ensuring a fair, consistent, and reproducible comparison across different models under the same test, as shown in Fig.~\ref{fig:eval-pos} for the example of the \textbf{\textit{Transport}} task.

\end{document}